\documentclass[letterpaper]{article} 
\usepackage{aaai2026}  
\usepackage{times}  
\usepackage{helvet}  
\usepackage{courier}  
\usepackage[hyphens]{url}  
\usepackage{graphicx} 
\usepackage{natbib}  
\usepackage{caption} 
\usepackage{algorithm}
\usepackage{algorithmic}

\usepackage{booktabs}
\usepackage{amsmath}
\usepackage{amsfonts}
\usepackage{multirow}
\usepackage{xcolor}
\usepackage{colortbl}
\definecolor{lightgray}{gray}{0.9}

\usepackage{newfloat}
\usepackage{listings}
\DeclareCaptionStyle{ruled}{labelfont=normalfont,labelsep=colon,strut=off} 
\floatstyle{ruled}
\newfloat{listing}{tb}{lst}{}
\floatname{listing}{Listing}
\title{From Dataset to Real-world: General 3D Object Detection via Generalized Cross-domain Few-shot Learning}
\author{
    Shuangzhi Li\textsuperscript{\rm 1}, Junlong Shen\textsuperscript{\rm 1}, Lei Ma\textsuperscript{\rm 2,1}, and Xingyu Li\textsuperscript{\rm 1}
}
\affiliations{
    \textsuperscript{\rm 1}University of Alberta, Canada \\ \textsuperscript{\rm 2}The University of Tokyo, Japan\\
    \{shuangzh, junlong6, xingyu\}@ualberta.ca, ma.lei@acm.org
}

\begin{document}

\maketitle

\begin{abstract}
LiDAR-based 3D object detection models often struggle to generalize to real-world environments due to limited object diversity in existing datasets. To tackle it, we introduce the first generalized cross-domain few-shot (GCFS) task in 3D object detection, aiming to adapt a source-pretrained model to both common and novel classes in a new domain with only few-shot annotations. We propose a unified framework that learns stable target semantics under limited supervision by bridging 2D open-set semantics with 3D spatial reasoning. Specifically, an image-guided multi-modal fusion injects transferable 2D semantic cues into the 3D pipeline via vision-language models, while a physically-aware box search enhances 2D-to-3D alignment via LiDAR priors. To capture class-specific semantics from sparse data, we further introduce contrastive-enhanced prototype learning, which encodes few-shot instances into discriminative semantic anchors and stabilizes representation learning. Extensive experiments on GCFS benchmarks demonstrate the effectiveness and generality of our approach in realistic deployment settings. 
\end{abstract}

\begin{links}
    \link{Code}{https://github.com/Castiel-Lee/GCFS-3Det}
\end{links}

\section{Introduction}
\label{sec:intro}

LiDAR-based 3D object detection~\cite{zhang2025new,baur2024liso,mao20233d} has significantly advanced autonomous driving by leveraging annotated datasets collected across diverse global locations~\cite{geiger2012we, caesar2020nuscenes, sun2020scalability, geyer2020a2d2}. However, as summarized in Table~\ref{tab: datasets_classes}, existing datasets primarily focus on a limited set of common object categories (such as cars, pedestrians, and bicycles) within selected urban areas (e.g., USA, Singapore, and German cities). In contrast, real-world deployment introduces new geographic regions and novel object categories, such as electric scooters in Chinese cities or tuk-tuks in Thailand. Collecting and annotating large-scale LiDAR datasets for each new environment is both time-consuming and resource-prohibitive, which makes it unsuitable for rapid adaptation. This practical limitation highlights the need for methods that can generalize beyond the constraints of existing datasets: adapting to new domains and emerging object categories with minimal supervision.

Despite growing interest in these challenges, existing LiDAR-based 3D detection methods still face key limitations in effectively generalizing to novel categories with limited target-domain data. Among existing approaches, semi-supervised learning~\cite{wang2023ssda3d} and 3D open-vocabulary detection (OVD) \cite{etchegaray2024find,zhang2025opensight,cao2024coda} often assume the availability of large amounts of unlabeled target data, which isn't always feasible in model deployment. While 3D domain adaptation (DA) ~\cite{wang2020train,yang2022st3d++,hegde2024attentive} focuses on addressing domain shifts, it does not explicitly account for novel object categories unseen during training. Simply labeling novel objects as "others" is often insufficient in safety-critical scenarios where object-specific recognition is necessary for decision making.

\begin{figure}[t]
  \centering
   \includegraphics[width=0.90\linewidth]{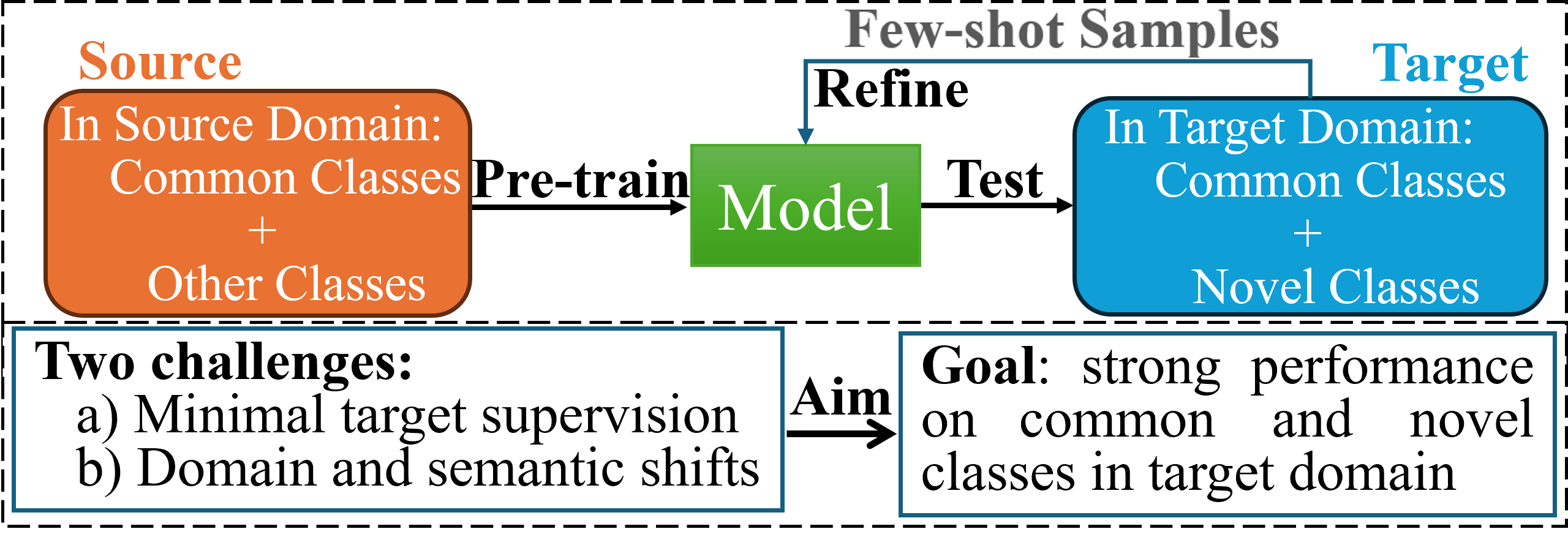}
   \caption{GCFS in 3D object detection aims to adapt source-pretrained models for strong performance on common and novel classes in the target domain via limited target samples. }
   \label{fig: highlight}
\end{figure}

\begin{table*}[]
\centering
{
\begin{tabular}{@{}cccll@{}}
\toprule
\textbf{Datasets}      & \textbf{Locations}         & \textbf{Classes} & \multicolumn{1}{c}{\textbf{Categories-of-interest}}                                                           &  \\ \midrule
KITTI~\shortcite{geiger2012we}        & Karlsruhe (Germany)         & $7$            & {\underline{Car, Pedestrian, Truck,}} Van, Person\_sitting, Cyclist, Tram                            &  \\
NuScenes~\shortcite{caesar2020nuscenes}     & Boston (USA), Singapore & $23$           & {\underline{Car, Pedestrian, Truck,}} Barrier, Construction\_vehicle, etc.    &  \\
Waymo~\shortcite{sun2020scalability}        & 3 cities in USA               & $4$            & Vehicle({\underline{car, truck,}} and bus), {\underline{Pedestrian,}} Cyclist, Sign                 &  \\
Argoverse 2~\shortcite{wilson2023argoverse} & 6 cities in USA               & $30$           & \underline{Car, Pedestrian, Truck}, Bicycle, Motorcycle, Bus, Barrel, etc.                   &  \\
A2D2~\shortcite{geyer2020a2d2}         & 50 cities in Germany           & $14$           & {\underline{Car, Pedestrian, Truck,}} Bicycle, Bus, UtilityVehicle, etc. &  \\ \bottomrule
\end{tabular}
}
\caption{Summary of common 3D Object Detection Datasets, where the most common detection categories are \underline{underlined}.}
\label{tab: datasets_classes}
\end{table*}

To bridge the gap from dataset-based training to real-world deployment, we tackle a new task, \textit{generalized cross-domain few-shot} (\textbf{GCFS}) learning, for LiDAR-based 3D object detection. 
As conceptualized in Fig.~\ref{fig: highlight}, the GCFS task comprehensively considers efficient adaptation to the target domain and stable semantic learning for novel and common categories via minimal target supervision, offering a cost-effective solution for rapid deployment in diverse environments.
Unlike existing 3D few-shot learning (FSL) ~\cite{zhao2022prototypical, tang2024prototypical, li2024cp}, or its extension, 3D generalized few-shot learning (GFSL)~\cite{liu2023generalized}, which assumes the same distribution between training and deployment environments, GCFS accommodates both domain discrepancies and semantic adaptation target under limited target supervision. 

Specifically, in GCFS tasks, a 3D object detection model is initially trained on a source dataset including common object classes along with other possible source-specific classes. In the target environment, which may have a domain gap from the source data due to environmental factors, sensor configurations, and object appearances~\cite{yang2022st3d++, hegde2024attentive, li2025domain}, we assume the presence of additional target-specific classes (i.e., novel classes) alongside the common classes. Given the practical feasibility and high cost of LiDAR data collection and annotation, we further assume that access to annotated data in the target environment is restricted to only a minimal amount (e.g., few-shot samples). The GCFS task, therefore, aims to enable the pre-trained model to adapt with minimal supervision in the target environment, ensuring strong performance on both common and target-specific novel categories. Although certain tasks in 2D object detection, such as few-shot domain adaptation~\cite{gao2023asyfod, nakamura2022few} and generalized few-shot learning~\cite{fan2021generalized,zhang2023generalized}, offer methodological insights into combining limited data adaptation with domain gap bridging, extending these 2D solutions effectively to the 3D domain remains challenging due to the higher-dimensional complexity and unique spatial characteristics of 3D data.

In this work, we introduce the first effective solution to comprehensively address the challenge of stable semantic representation learning under minimal target supervision in GCFS tasks. Our key insight is that generalization across domains and object categories is possible by bridging 2D open-set semantics and 3D spatial reasoning. By aligning sparse 3D observations with rich 2D vision-language priors and refining object understanding through prototype-based semantic anchoring, models can adapt robustly to both domain shifts and novel object classes from a few labeled examples. To realize this, we propose a unified GCFS framework built on two synergistic components: (1) an image-guided multi-modal fusion module that injects transferable 2D semantic cues into the 3D detection pipeline, improving proposal quality even in sparse point clouds; and (2) a contrastive-enhanced prototype learning mechanism that encodes few-shot target samples into discriminative, class-specific semantic anchors. Notably, we introduce a physically-aware box search strategy to improve 2D-to-3D alignment, and use contrastive learning to stabilize semantic prototypes under limited data. Together, these components enable robust adaptation with minimal supervision, offering a practical and generalizable solution for real-world 3D object detection. 
In evaluation, we design four GCFS benchmark settings and conduct extensive experiments to illustrate the effectiveness of our solution. 
In sum, our contributions are: 
\begin{itemize}
    \item We formulate the generalized cross-domain few-shot task for 3D object detection and propose the first GCFS solution, holistically addressing domain shifts and novel object categories under limited supervision. 

    \item We propose a unified framework that leverages image-guided semantic grounding and contrastive prototype refinement to learn transferable object-level representations from sparse 3D data. Our framework illustrates that combining 2D vision-language priors with 3D geometry and few-shot semantic anchoring enables robust generalization across diverse environments and categories.
    
    \item We establish four GCFS benchmark settings and show that our approach outperforms existing methods, providing a standardized framework for future research on 3D detection under domain and data constraints.
\end{itemize}

\section{Related Works}
\label{sec:related_works}

\subsection{LiDAR-based 3D Object Detection} 
LiDAR-based 3D object detection~\cite{zhang2025new, gambashidze2024weak,mao20233d} aims to locate and classify objects of interest from input point clouds. Its models are primarily categorized into point-based, voxel-based, and point-voxel-based methods. Point-based models~\cite{pan20213d,shi2019pointrcnn,shi2020point} incorporate raw points and the PointNet-based backbones for 
fine-grained representation at the point level, albeit with high computational demands. Voxel-based methods~\cite{yan2018second,mao2021voxel,deng2021voxel,zhou2018voxelnet} represent the point cloud within a structured voxel grid and utilize sparse convolution for feature extraction, offering a trade-off between computational efficiency and spatial resolution. Point-voxel-based methods~\cite{shi2023pv, shi2020pv} combine both, 
achieving a balance between efficiency and representation resolution, but often coming with increased model complexity and computation. 

\subsection{Few-shot Learning in Object Detection} 
In object detection, FSL aims to enable models to detect objects with limited labeled samples. 
In 2D, extensive studies~\cite{zhang2025few,xin2024few} tackle data scarcity by exploiting techniques like meta-learning~\cite{yan2019meta,app12115543}, transfer learning~\cite{wang2020frustratingly,chen2018lstd}, and data augmentation~\cite{wu2020multi}. 
In 3D object detection, most works focus on indoor scenarios.  
Based on VoteNet~\cite{qi2019deep}, 
Proto-Vote~\cite{zhao2022prototypical} introduces a prototypical vote module  
for local features refinement and a prototypical head module 
for global feature enhancement.
On top of it, a VAE-based prototype learning~\cite{tang2024prototypical} is designed, and
contrastive learning ~\cite{li2024cp} 
is further exploited  
to learn more refined prototypical representations. 
However, extending 3D indoor object detection methods to outdoor scenarios is challenging due to sparse point clouds at greater distances, dynamic objects, and varying lighting and weather conditions. 
A recent work ~\cite{liu2023generalized} proposes the first outdoor generalized FSL solution for novel class learning. Yet, without dealing with domain gaps in cross-domain scenarios, it leads to limited performance on GCFS settings.

\subsection{Domain Adaptation in 3D Object Detection}
The study of domain adaptation in 3D object detection mainly focuses on unsupervised or semi-supervised settings. 
Works~\cite{yang2022st3d++,chen2018lstd} employ a hybrid quality-aware triplet memory to generate pseudo-labels for unlabeled target-domain data. A source-free unsupervised DA approach~\cite{hegde2024attentive} utilizes class prototypes to suppress noisy pseudo-labels on target data. 
Density-resampling-based augmentation and test-time adaptation~\cite{li2025domain} are proposed to bridge density-related domain gaps.
Yet, dependence on large target datasets and the inability to handle novel classes make these methods inapplicable to GCFS tasks

\subsection{Open-vocabulary 3D Object Detection}
Recently, open-vocabulary object detection~\cite{wu2024towards, zareian2021open, Gu2021OpenvocabularyOD, zhang2023simple, li2021grounded} has garnered significant attention. In 3D object detection, these methods usually take advantage of 2D VLMs to acquire novel open-set semantics and enable detection on novel objects without annotations. 
For instance, \citeauthor{lu2023open} \citeyear{lu2023open} proposes to utilize CLIP-based VLMs to connect open-set textual knowledge and point-cloud representations for novel object identification. 
Auto-label methods ~\cite{najibi2023unsupervised, etchegaray2024find} are applied to point cloud sequences via a pretrained 2D VLM and enable novel semantic discovery for self-training.
A 2D-3D co-modeling approach~\cite{zhang2025opensight} estimates corresponding 3D boxes from 2D insights with temporal and spatial constraints. Since these 3D-OVD methods rely on large volumes of target data (including novel objects), their performance on the GCFS task remains to be validated.


\begin{figure*}[t]
  \centering
   \includegraphics[width=0.84\linewidth]{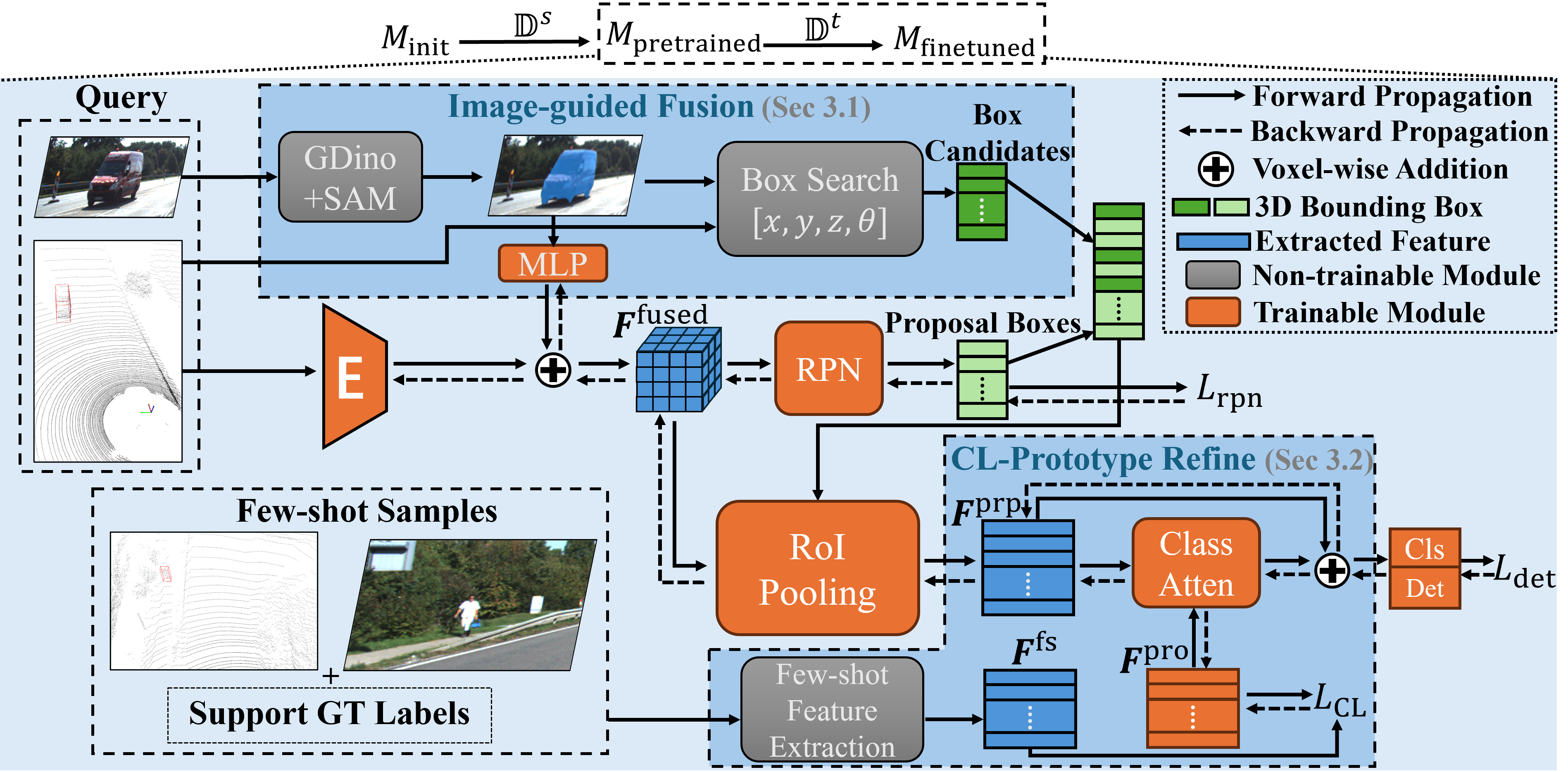}
   \caption{Proposed GCFS Framework. 
   We first pretrain a detection model with source data. During model finetuning using target few-shot samples, each query—the image and point cloud pair—is processed by GDino+SAM and 3D backbone for 2D instance-level masks and 3D features (top block). Insights from 2D context contribute to 1) enriching 3D features $F^{\text{fused}}$ with 2D semantic clues and 2) proposing high-quality ``Box Candidates'' via a novel 2D-to-3D box search. Proposal features $\textbf{F}^{\text{prp}}$ are refined by learnable prototypes $\textbf{F}^{\text{pro}}$ with an attention mechanism, and then passed to the final prediction (bottom block).
   }
   \label{fig: frame_work_overview}
\end{figure*}

\section{Methodology}
\label{sec:methodology}
\textbf{Problem Statement:} To formulate the GCFS of 3D object detection, we distinguish LiDAR data from the source dataset and target environment (dataset) with superscripts $s$ and $t$, respectively. In the source dataset used to pre-train the model $M_{\text{pretrained}}$, 
we assume access to sufficient annotated data $\mathbb{D}^s=\{\textbf{B}^s_i,\textbf{C}^s_i,\textbf{P}^s_i\}_{i=1}^{N^s}$, where $\textbf{P}^s_i \in \mathbb{R}^{N{\text{pts}} \times 3}$ denotes the point cloud, $\textbf{B}^s_i = \{\textbf{b}^s\mid\textbf{b}^s=[x,y,z,h,w,l,\theta]\}_{l=1}^{N_{\text{obj}}}$ the 3D bounding boxes, and $\textbf{C}^s_i$ the corresponding object category belonging to the source category space $\mathbb{C}^s$. For the target dataset $\mathbb{D}^t = {(\textbf{B}^t_i, \textbf{C}^t_i, \textbf{P}^t_i)}_{i=1}^{N^t}$, only limited (few-shot) samples are available for each target object category in the target category set $\mathbb{C}^t$.
Here, we assume some categories are shared in $\mathbb{C}^t$ and $\mathbb{C}^s$, so certain knowledge in $M_{\text{pretrained}}$ is valuable to the target task. Formally, these common classes are defined by $\mathbb{C}_\text{com}=\mathbb{C}^t\cap\mathbb{C}^s\neq\emptyset$. We use $\mathbb{C}_\text{nov}^s=\mathbb{C}^s\setminus\mathbb{C}_\text{com}$ and $\mathbb{C}_\text{nov}^t=\mathbb{C}^t \setminus \mathbb{C}_\text{com}$ to denote the domain-specific novel classes. That is, objects belonging to $\mathbb{C}_\text{nov}^t$ are unseen in the source dataset. The goal of GCFS tasks is to obtain a strong detection model $M_{\text{finetuned}}$ through refining $M_{\text{pretrained}}$ with the $K$-shot examples in $\mathbb{D}^t$.%

Fig.~\ref{fig: frame_work_overview} presents an overview of our framework. 
To learn stable target semantics under limited supervision, we integrate two key components: an \textit{image-guided multi-modal fusion} module and a \textit{class-specific contrastive prototype learning} module. The fusion module exploits vision-language models (VLMs) to extract open-set semantic cues from point-cloud-aligned images, guided by a physically-aware box searching strategy that models LiDAR scanning behavior in the 3D geometric space. Meanwhile, the prototype learning module encodes class-level semantics from few-shot target samples into discriminative prototype anchors, which refine and align object features during inference. 

\subsection{Image-guided Multi-modal Fusion (IMMF)}
\label{sec:method_image_guided_fusion}

In GCFS tasks, detectors trained on source data must adapt to new domains and categories via minimal target supervision. Yet, LiDAR data, which is sparse and geometry-focused, offers limited semantic richness, especially for novel objects. In contrast, aligned RGB images offer dense, transferable visual features and access to open-set semantics via pre-trained VLMs. To bridge this semantic gap, we introduce an image-guided multi-modal fusion that enriches 3D point representations with 2D semantic cues extracted from Grounding DINO (GDino)~\cite{liu2024grounding} and SAM~\cite{kirillov2023segment}, improving detection robustness under domain and category shifts.

\noindent \textbf{Image-guided feature fusion.}
Given the point cloud $\textbf{P}$, we extract the non-empty voxel feature $\textbf{F}^{\text{voxel}} \in \mathbb{R}^{N_{\text{voxel}} \times C}$ via a 3D backbone, where $C$ denotes the 3D feature dimension. For the paired image $\textbf{I} \in \mathbb{R}^{H \times W \times 3}$, we use object category names (i.e.,  $\mathbb{C}_\text{com}$ and $\mathbb{C}_\text{nov}^t$) as text prompts to activate GDino, producing $N_\text{obj}$ 2D boxes $\textbf{B}^\text{2D} \in \mathbb{R}^{N_\text{obj} \times 4}$ with class labels as potential semantic clues. After non-maximum suppression and confidence filtering, SAM takes $\textbf{B}^\text{2D}$ as box prompts and generates dense object masks $\textbf{M}^\text{2D} \in \mathbb{R}^{H \times W \times |\mathbb{C}^{t}|}$. We then project the coordinates $\textbf{P}^\text{voxel}$ of $\textbf{F}^\text{voxel}$ onto the image to identify the object masks and obtain the voxel-aligned object mask $\textbf{M}^\text{vxl-obj} \in \mathbb{R}^{N_\text{voxel} \times |\mathbb{C}^{t}|}$:
\begin{equation}
\textbf{M}^\text{vxl-obj} = f_\text{proj}(\textbf{M}^\text{2D}, \textbf{P}^\text{voxel}),
\end{equation}
where $f_\text{proj}(\cdot)$ denotes 2D-to-3D mapping based on known camera intrinsics and extrinsics.
To integrate 2D semantic cues into the 3D representation, we apply an MLP to align the channel dimensions and fuse the features:
\begin{equation}
\textbf{F}^\text{fused} = \textbf{F}^\text{voxel} + \text{MLP}(\textbf{M}^\text{vxl-obj}).
\end{equation}
This fused feature $\textbf{F}^\text{fused}$ enhances the downstream region proposal network (RPN), improving object recall for both common and novel categories.

\begin{figure}[t]
  \centering
   \includegraphics[width=0.96\linewidth]{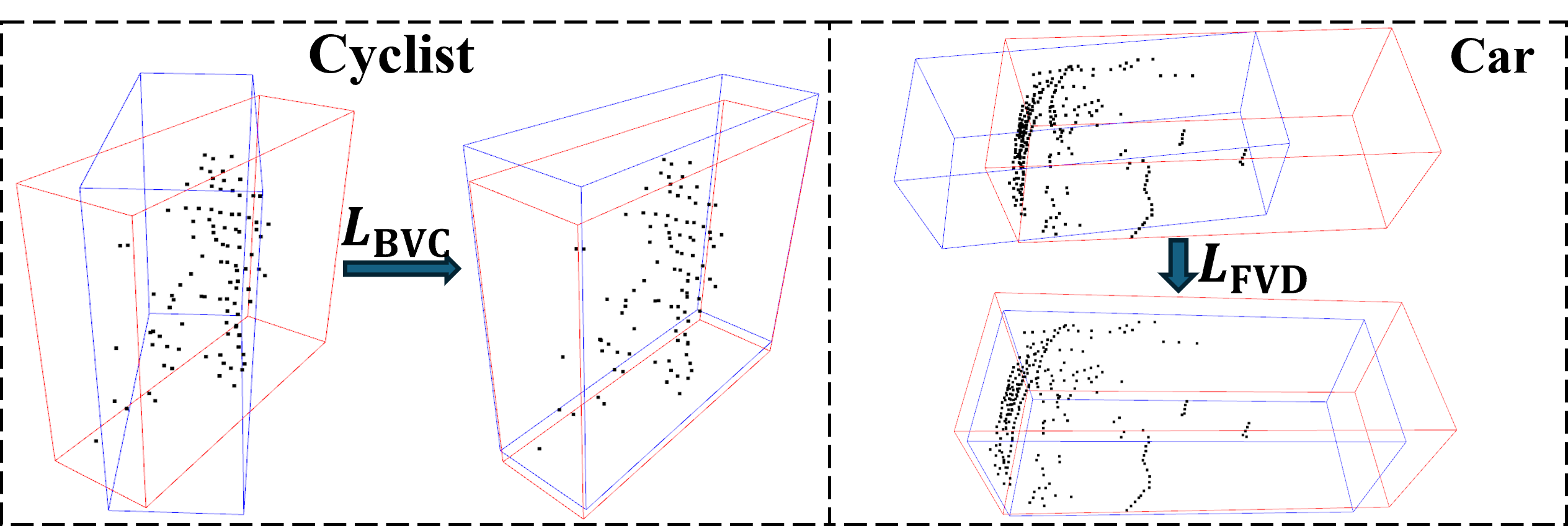}
   \caption{Physical-aware box searching. Red boxes are GT boxes, and blue ones are searched boxes.
   Regarding ``Cyclist'' (left) and ``Car'' (right), angle and center biases on searched boxes are corrected by $L_\text{BVC}$ and $L_\text{FVD}$.
   }
   \label{fig: box_search}
\end{figure}

\noindent \textbf{Physical-aware 3D box searching from 2D masks.} While VLMs provide rich semantics, transferring these 2D cues into the 3D space is inherently noisy in sparse LiDAR settings. Calibration inaccuracies and vision misalignment can lead to imprecise 2D-to-3D mappings. To ensure 2D semantic cues are projected to geometrically plausible 3D box proposals, we introduce a physically-aware box search strategy that filters and aligns proposals based on spatial consistency. 

Specifically, to estimate fine-grained box locations from the 2D object masks $\textbf{M}^\text{2D}$, we first project the raw point cloud $\textbf{P}$ into the image and identify points within masks by $\textbf{P}^\text{pts} = f_\text{proj}(\textbf{M}^\text{2D}, \textbf{P})^\top\textbf{P}$, where $\textbf{P}^{\text{pts}}$ denotes the points of all object masks.
For the $i^{th}$ object, we extract its points $\textbf{P}^\text{pts}_{i}\in\textbf{P}^\text{pts}$ and use the mean and $2\times$standard deviation of point coordinates as the center and boundary of the valid range to eliminate background points. For each class $c \in \mathbb{C}^t$, we pre-define an anchor box with the size $[h^c, w^c, l^c]$ via the mean size of target few-shot objects. The goal of box searching is to find the optimal center $[x, y, z]$ and heading angle $\theta$ of the anchor box for each object. Specifically, for $i$-th object, $[x, y, z, \theta]$ defines a rotation transformation $\textbf{T}$ (see the supplementary for details), and centered coordinates $\textbf{P}^\text{local}_{i}$ are obtained by $\textbf{P}^\text{local}_{i}=\textbf{T} \textbf{P}^\text{pts}_{i}$.
We first design an outside distance loss $L_\text{OD}$ to constrain $\textbf{P}^\text{local}_{i}$ in the box, 
\begin{equation}
    L_{OD} = \sum_{\textbf{p} \in \textbf{P}^\text{local}_{i}} \textrm{min}(\textrm{abs}(\textbf{p}) - \textbf{BD}^c, \textbf{0}).
\end{equation}
Here, $\textbf{BD}^c = [h^c/2, w^c/2, l^c/2]$ denotes the local box boundary for class $c$. 

Furthermore, we notice that, due to central unidirectional scanning, LiDAR-scanned object points present significant differences in point distribution regarding different structural complexities. For simple structural objects with flat surfaces, like vehicles (e.g., cars and buses), most points are on smooth surfaces and front-viewed by LiDAR. For complex structural objects with irregular surfaces (e.g., pedestrians and bicycles),  points are more to shape the whole objects in the bird's eye view. Motivated by this observation, we categorize general objects into two types: simple structural (SS) objects and complex structural (CS) ones. 
For SS objects, we design the front-viewed distance (FVD) loss to make points closer to the front-viewed boundaries of the box,
\begin{equation}
    L_\text{FVD} = \sum_{\textbf{p} \in \textbf{P}^\text{local}_{i}} ||\textbf{p} - \textbf{FB}^c||\cdot\mathbf{1}(\textbf{P}^\text{local}_{i}\in \text{SS}),
\end{equation}
where the LiDAR front-viewed box boundaries $\textbf{FB}^c$ is defined by $[x, y, z, \theta]$ (see the supplementary for details).
For CS objects, we design the bird-viewed center (BVC) loss to align the centers of points and boxes. 
\begin{equation}
    L_\text{BVC} = \sum_{\textbf{p} \in \textbf{P}^{local}_{i}} ||f_{2D}(\textbf{p})||\cdot\mathbf{1}(\textbf{P}^{local}_{i}\in \text{CS}),
\end{equation}
where $f_\text{2D}(\cdot)$ simply obtains $[x, y]$ of points.
Applying $L_\text{BVC}$ and $L_\text{FVD}$ facilitates the discovery of the correct centers and heading angles for boxes, as shown in Fig.~\ref{fig: box_search}. 
In summary, the box-searching loss for optimizing $[x, y, z, \theta]$ is:
\begin{equation}
    L_\text{box} = L_\text{OD} + \lambda_{1}L_\text{FVD} + \lambda_{2}L_\text{BVC}.
\end{equation} 
Since the computational load of box searching is low (due to sparse object points), we use the Quasi-Newton BFGS optimization~\cite{head1985broyden} to efficiently optimize $[x,y,z, \theta]$ for each object. In essence, our physically-aware box search acts as a semantic gatekeeper-ensuring that 2D-to-3D knowledge transfer remains spatially coherent.

\subsection{Class-specific Contrastive-Enhanced Learnable Prototype and Feature Refinement}
\label{sec:method_meta_learning} 

While our IMMF module improves proposal accuracy, domain shifts and limited annotations still hinder reliable feature learning via simple fine-tuning. To overcome this, we propose a contrastive prototype learning strategy that builds robust, class-specific semantic anchors from limited examples and enhances them using contrastive learning to increase generalization and inter-class separability.
Unlike the work \cite{li2024cp}, which uses contrastive learning to enhance static prototypes, our approach uses few-shot-driven contrastive learning on learnable prototypes, making our prototypes more discriminative.

\begin{figure}[t]
  \centering
   \includegraphics[width=0.90\linewidth]{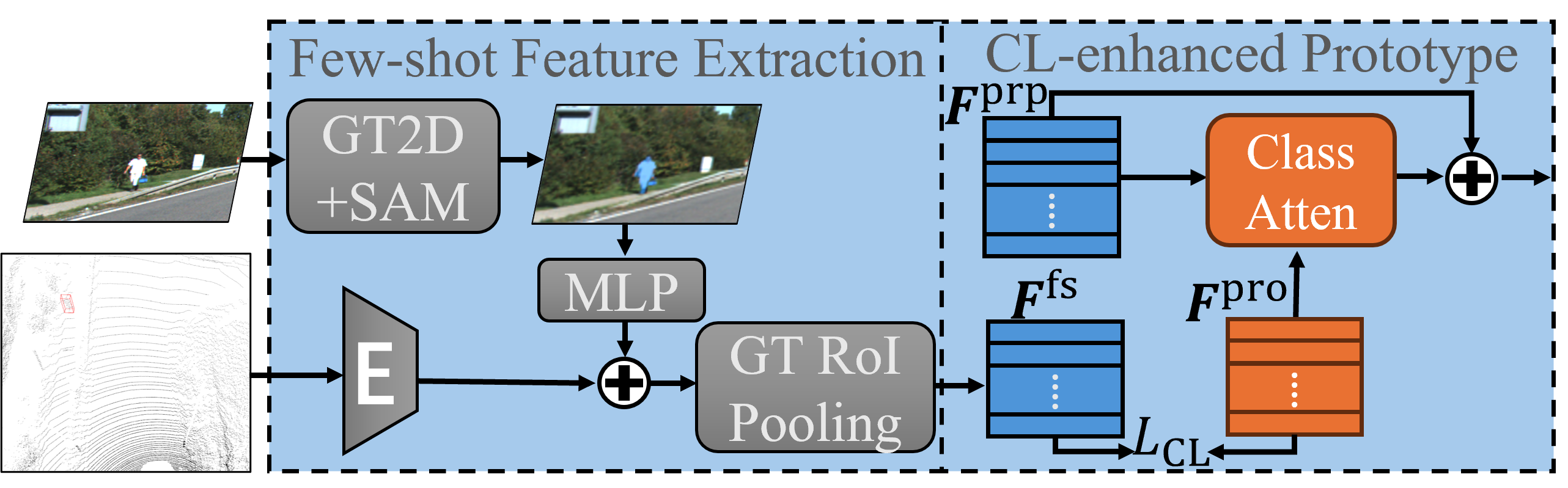}
   \caption{Few-shot feature extraction and CL-enhanced prototype learning. In few-shot feature extraction, 2D and 3D ground-truth labels replace GDino and RPN outputs to extract object features.}
   \label{fig: proto_refine}
\end{figure}

\noindent \textbf{Class-specific contrastive prototype learning.} 
We build a learnable target-specific feature bank $\textbf{F}^\text{pro} \in \mathbb{R}^{|\mathbb{C}^{t}| \times d}$ for all object classes, where $d$ is the dimension of features. These prototypes are optimized together with the model fine-tuning update. To accelerate convergence under limited data, we introduce a contrastive loss for the learnable prototypes. As shown in Fig.~\ref{fig: proto_refine}, we group the features of the few shots $\textbf{F}^\text{fs}$ according to their box annotation as contrastive anchors. Then for each class $c\in \mathbb{C}^t$, we construct positive pairs with the corresponding prototype $\textbf{F}^\text{pro}_c$ and its anchor $\textbf{F}^\text{fs}_c$. The remaining prototypes in the feature bank, denoted by $\textbf{F}^\text{pro}_s$, are negative samples of the anchor. 
\begin{equation}
    L_\text{CL} = -\sum_{c \in \mathbb{C}^{t}}\log\frac{\exp(\text{Sim}(\textbf{F}^\text{fs}_c, \textbf{F}^\text{pro}_c)/\tau)}{\sum_{s \in \mathbb{C}^{t}}\exp(\text{Sim}(\textbf{F}^\text{fs}_c, \textbf{F}^\text{pro}_s)\tau)},
\end{equation}
where $\text{Sim}(\cdot,\cdot)$ calculates the cosine similarity between two features in the InfoNCE loss~\cite{oord2018representation} with a temperature $\tau$. Since the anchors are directly obtained from target-domain examples, our contrastive-enhanced features help bridge the domain gap between source and target environments and speed up $\textbf{F}^\text{pro}$ acquiring semantic essences of various classes under limited training data.

\noindent \textbf{Feature refinement by prototypes.}
After obtaining the $\textbf{F}^\text{pro}$ along with the model finetuning process, we use them to refine the proposal features $\textbf{F}^\text{prp}$ of the query input. In the multi-head cross-attention, we take $\textbf{F}^\text{pro}$ to form the key and value, and $\textbf{F}^\text{prp}$ as the query. 
\begin{equation}
    \hat{\textbf{F}}^\text{prp}=\text{Softmax}(\frac{{\textbf{F}}^\text{prp}\textbf{W}_\text{Q}({\textbf{F}}^\text{pro}\textbf{W}_\text{K})^\top}{\sqrt{d}}){\textbf{F}}^\text{pro}\textbf{W}_\text{V},
\end{equation}
where [$\textbf{W}_\text{Q}, \textbf{W}_\text{K}, \textbf{W}_\text{V}$] is the trainable transformation of the query, key, and value. Finally,  
\begin{equation}
\tilde{\textbf{F}}^\text{prp}=\hat{\textbf{F}}^\text{prp}+\textbf{F}^\text{prp},
\end{equation}
is passed to the object detection head for object detection. 

\subsection{Model Optimization and Inference} The model parameter update and our prototype learning are conducted together. The Overall loss to optimize them is:
\begin{equation}
    L=L_\text{rpn}+L_\text{det}+{\lambda}L_\text{CL},
\end{equation}
where $L_\text{rpn}$ and $L_\text{det}$ are the standard losses of RPN and detection head, and $\lambda$ is a weight hyper-parameter. 
To further enable the model's adaptability to a new domain under limited data, we adopt an MAML-based~\cite{finn2017model} training scheme. 
Briefly, during meta-training, we leverage the source data to set up the $K$-shot meta-task.
This meta-training facilitates finding a set of model parameters and $\textbf{F}^\text{pro}$ for the quick model adaptation in the unseen domain (see the supplementary for more details). During deployment, aligned point clouds and images undergo the proposed image-guided fusion to enhance semantic discovery in proposals. After ROI pooling, object features are further refined with class prototypes to improve discrimination.

\begin{table*}[t]
\centering
\setlength{\tabcolsep}{1.3pt}
{%
\begin{tabular}{@{}c|ccc|ccc|ccc|ccc@{}}
\toprule
\multirow{2}{*}{\textbf{Methods}} & \multicolumn{3}{c|}{\textbf{NuScenes→FS-KITTI}} & \multicolumn{3}{c|}{\textbf{Waymo→FS-KITTI}} & \multicolumn{3}{c|}{\textbf{KITTI→FS-A2D2}} &  \multicolumn{3}{c}{\textbf{KITTI→FS-Argo2}} \\ \cmidrule(l){2-13} 
&  \textbf{common} & \textbf{novel} & \textbf{overall} & \textbf{common} & \textbf{novel} & \textbf{overall} & \textbf{common} & \textbf{novel} & \textbf{overall} & \textbf{common} & \textbf{novel} & \textbf{overall} \\ \midrule
Source-only & 14.24 & - & - & \textbf{26.85} & - & - & 3.81 & - & - & 6.65 & - & - \\
Target-FT & 12.77\textsubscript{(1.9)} & 5.48\textsubscript{(1.2)} & 8.61\textsubscript{(1.5)} & 23.06\textsubscript{(1.6)} & 12.47\textsubscript{(1.9)} & 17.01\textsubscript{(1.8)} & 5.09\textsubscript{(1.1)} & 0.70\textsubscript{(0.2)} & 2.90\textsubscript{(0.6)} & 3.18\textsubscript{(0.2)}  & 0.62\textsubscript{(0.1)} & 1.39\textsubscript{(0.1)} \\
Proto-Vote & 7.56\textsubscript{(2.4)} & 5.74\textsubscript{(1.5)} & 6.52\textsubscript{(1.9)} & 17.36\textsubscript{(2.7)} & 12.08\textsubscript{(1.8)} & 14.34\textsubscript{(2.2)} & 3.61\textsubscript{(0.8)} & 1.86\textsubscript{(0.5)} & 2.74\textsubscript{(0.7)} &  3.33\textsubscript{(0.9)}   & 0.90\textsubscript{(0.4)}  & 1.63\textsubscript{(0.5)} \\
PVAE-Vote & 8.01\textsubscript{(2.8)} & 6.38\textsubscript{(2.2)} & 7.08\textsubscript{(2.5)} & 18.19\textsubscript{(2.9)} & 12.79\textsubscript{(2.2)} & 15.10\textsubscript{(2.5)} & 3.43\textsubscript{(0.9)} & 1.97\textsubscript{(0.5)} & 2.70\textsubscript{(0.7)} & 3.10\textsubscript{(1.0)}   & 0.92\textsubscript{(0.3)} & 1.58\textsubscript{(0.5)}  \\
CP-Vote & 10.69\textsubscript{(2.3)} & 7.84\textsubscript{(1.9)} & 9.06\textsubscript{(2.0)} & 17.66\textsubscript{(2.4)} & 12.17\textsubscript{(1.9)} & 14.52\textsubscript{(2.1)} & 4.28\textsubscript{(0.9)} & 2.72\textsubscript{(0.9)} & 3.50\textsubscript{(0.9)} & 2.72\textsubscript{(0.9)}  & 0.93\textsubscript{(0.4)} & 1.47\textsubscript{(0.5)} \\
GFS-Det & 12.83\textsubscript{(2.4)} & 1.18\textsubscript{(0.4)} & 6.17\textsubscript{(1.2)} & 22.74\textsubscript{(2.8)} & 1.26\textsubscript{(0.4)} & 10.47\textsubscript{(1.4)} & 4.39\textsubscript{(0.6)} & 0.22\textsubscript{(0.1)} & 2.30\textsubscript{(0.3)} & 6.11\textsubscript{(0.1)}  & 0.03\textsubscript{(0.0)} & 1.86\textsubscript{(0.0)} \\ \midrule 
\rowcolor{lightgray}\textbf{Ours} & \textbf{15.99}\textsubscript{(1.6)} & \textbf{11.72}\textsubscript{(1.4)} & \textbf{13.55}\textsubscript{(1.5)} & 25.40\textsubscript{(2.0)} & \textbf{17.75}\textsubscript{(1.7)} & \textbf{21.03}\textsubscript{(1.8)} & \textbf{7.78}\textsubscript{(0.7)} & \textbf{5.22}\textsubscript{(0.6)} & \textbf{6.50}\textsubscript{(0.6)}  & \textbf{6.71}\textsubscript{(0.2)}  & \textbf{2.07}\textsubscript{(0.2)} & \textbf{3.46}\textsubscript{(0.2)}\\  \midrule
Full-Target & 41.34 & 18.35 & 28.21 & 41.34 & 18.35 & 28.21 & 36.61 & 5.99 & 21.30 & 31.75 & 18.48 & 22.46 \\
 \bottomrule
\end{tabular}%
}
\caption{Performance in mAP(\%) of VoxelRCNN for NuScenes $\rightarrow5$shot-KITTI, Waymo $\rightarrow5$shot-KITTI, KITTI $\rightarrow5$shot-A2D2, and KITTI $\rightarrow5$shot-Argo2. 
The \textbf{bold} values represent the best performance except Full-Target. Subscript values in parentheses are standard deviations. Please refer to the supplementary for specifics across various categories.}
\label{tab:main_3GCFS_two_models}
\end{table*}

\section{Experimentation}
\label{sec:experiment}

\subsection{Experimental Settings\protect\footnote{Details on the benchmark setup and implementation are provided in the supplementary linked in our GitHub repository.}}
\label{sec:exp_set}
\textbf{Benchmarks.} Since no prior study on GCFS tasks in 3D object detection, we leverage Nuscenes~\shortcite{caesar2020nuscenes}, Waymo~\shortcite{sun2020scalability}, KITTI~\shortcite{geiger2012we}, A2D2~\shortcite{geyer2020a2d2}, and Argoverse 2~\shortcite{wilson2023argoverse} to construct 4 GCFS benchmarks: NuScenes$\rightarrow$FS-KITTI, Waymo$\rightarrow$ FS-KITTI, KITTI$\rightarrow$FS-A2D2, and KITTI$\rightarrow$FS-Argo2. Specifically, we construct few-shot datasets by sampling K-shot objects per class from the \textit{train} set of KITTI, A2D2, and Argoverse 2, forming FS-KITTI, FS-A2D2, FS-Argo2. We set $K=5$ for main experiments, while our ablation study explores $K\in \{1,3,5,10, 20, 40\}$ for a comprehensive evaluation. The \textit{val} sets of KITTI and Argoverse 2 and the \textit{test} set of A2D2 are used for model evaluation. 
According to Table~\ref{tab: datasets_classes}, we select [\textit{Car, Pedestrian, Truck}] as common classes for all datasets. 
For sufficient samples for model evaluation and avoiding class ambiguity, we target novel classes: [\textit{Van, Person\_sitting, Cyclist, Tram}] in  FS-KITTI, [\textit{Bicycle, Utility\_vehicle, Bus}] in FS-A2D2, and [\textit{Construction\_barrel, Traffic\_cone, Large\_vehicle, Bicycle, Bus, Motorcycle, Sign}] in FS-Argo2. We use Average Precision (AP) to measure precision-recall trade-offs for each class~\cite{geiger2012we} and mean Average Precision (mAP) across multi-classes to assess overall performance. We conduct experiments 5 times and report the average mAP across trials, along with the standard deviation for stability evaluation. 

\noindent \textbf{Implementation Details.}  
We use VoxelRCNN~\cite{deng2021voxel} (voxel-based) and PV-RCNN++~\cite{shi2023pv} (point-voxel-based) as base detectors. Pre-training applies standard augmentations: random world flipping, scaling, and rotation. In fine-tuning, we additionally use ground-truth object sampling to ensure all target classes are present in each iteration. For box searching, we define SS classes [\textit{Car, Truck, Van, Tram, Bus, Construction\_barrel, Large\_vehicle, Sign}] and CS classes [\textit{Pedestrian, Person\_sitting, Cyclist, Bicycle, Utility\_vehicle, Traffic\_cone, Motorcycle}].
The Adam-OneCycle optimizer~\cite{openpcdet2020,song2024robofusion} is used with a 0.01 learning rate. All models are pre-trained for 30 epochs on NuScenes and Waymo, 80 epochs on KITTI, and fine-tuned for 100 epochs in FS-datasets.  
Batch sizes are 2 in pre-training and 1 in fine-tuning and testing.

\noindent \textbf{Compared Methods.} As no prior work has specifically tackled GCFS tasks for outdoor 3D object detection, we use a simple fine-tuning on few-shot target data (Target-FT) as the baseline. 
Source-only training and full target supervision (Source-only and Full-Target) serve as the performance with no and full adaptation.
To benchmark our method, we compare against SOTA 3D-FSL methods, Proto-Vote~\cite{zhao2022prototypical}, PVAE-Vote~\cite{tang2024prototypical}, and CP-Vote~\cite{li2024cp}, as well as the 3D-GFSL method GFS-Det~\cite{liu2023generalized}. Note that current outdoor OVD methods (i.e., Unsup3D~\cite{najibi2023unsupervised}, FnP~\cite{etchegaray2024find}, and OpenSight~\cite{zhang2025opensight}) and 3D-DA methods (i.e., SN~\cite{wang2020train}, ST3D++~\cite{yang2022st3d++,yang2021st3d}, and DenResamp~\cite{li2025domain}) are not directly applicable to our GCFS benchmark, as they rely on extensive unannotated data for unsupervised learning.
To further assess the generalizability and potential of our approach, we extend our ablation study to a more complex unsupervised few-shot learning setting, where these 3D-OVD and 3D-DA methods can be evaluated under conditions more aligned with their original assumptions.

\begin{table}[t]
\centering
\setlength{\tabcolsep}{2pt}
{%
\begin{tabular}{@{}c|ccc|ccc@{}}
\toprule
 & \begin{tabular}[c]{@{}c@{}}\textbf{Target-}\\ \textbf{FT}\end{tabular} & \begin{tabular}[c]{@{}c@{}}\textbf{Image-}\\ \textbf{Fusion}\end{tabular} & \begin{tabular}[c]{@{}c@{}}\textbf{CL-}\\ \textbf{Proto}\end{tabular} & \textbf{Common} & \textbf{Novel} & \textbf{Overall} \\ \midrule
(a) & \checkmark &  &  & 12.77 & 5.48 & 8.61 \\
(b) & \checkmark &  & \checkmark & 14.80 & 8.10 & 10.97 \\
(c) & \checkmark & \checkmark &  & 14.69 & 11.17 & 12.68 \\ \midrule
(d) & \checkmark & \checkmark & \checkmark & \textbf{15.99} & \textbf{11.72} & \textbf{13.55} \\ \bottomrule
\end{tabular}
}
\caption{Component ablations in mAP(\%). \textbf{Image-Fusion} is our proposed IMMF module and \textbf{CL-Proto} is our proposed contrastive-learning-enhanced prototype learning.}
\label{tab:component_ablation}
\end{table}

\subsection{Experimental Results on GCFS Benchmark}
\label{sec:main_exp} 

As shown in Table~\ref{tab:main_3GCFS_two_models}, our method consistently achieves superior performance in all GCFS benchmarks, demonstrating strong generalization to both common and novel categories under limited supervision. It arises from two key strengths. First, our method exhibits robust cross-domain transferability under diverse density-domain shifts, including varying LiDAR configurations across NuScenes (32-beam), Waymo (64-beam), KITTI (64-beam), A2D2 (16-beam), and Argoverse 2 (32-beam). It effectively maintains detection quality despite drastic variations in point density and sensor characteristics. Second, our approach enables efficient few-shot adaptation to target semantic concepts, as evidenced by its performance in semantically challenging settings like KITTI$\rightarrow5$shot-Argo2, involving seven diverse novel classes. In contrast, 3D-FSL methods show limited robustness on common classes due to their reliance on dense, close-range point clouds. Meanwhile, GFSL-Det struggles to generalize to novel classes, as its simplistic incremental learning strategy lacks mechanisms for semantic transfer from common classes to novel ones.

\noindent \textbf{Limitations and Future Work.}
Our method shows limited gains on certain hard classes (e.g. ``Person\_sitting'') due to ambiguous and diverse structures. In low-shift scenarios without semantic changes (Waymo $\rightarrow$ FS-KITTI), improvements on common classes are marginal, due to the interruption of novel classes. Future work will focus on hard class learning, adaptability in shiftless settings, and code optimization for computation speed-up.

\subsection{Ablation Studies}
\label{sec:ablation_studies}

We conduct ablation experiments mainly on NuScenes $\rightarrow5$shot-KITTI with VoxelRCNN as detection model, to further analyze our method (see the supplementary for details).

\begin{table}[t]
  \begin{minipage}{0.48\linewidth}
    \centering
    \setlength{\tabcolsep}{2pt}
    {%
    \begin{tabular}{@{}ccc@{}}
    \toprule
    \textbf{Prototype} & \textbf{Common} & \textbf{Novel} \\ \midrule
    w/o CL & 15.23 & 10.33 \\
    w/ CL & \textbf{15.99} & \textbf{11.72} \\ \bottomrule
    \end{tabular}%
    \caption{Performance in mAP(\%) of prototype learning with or without contrastive learning (CL).}
    \label{tab:ablation_CL_proto}
    }
  \end{minipage}
  \hspace{0.03\linewidth}
  \begin{minipage}{0.47\linewidth}
    \centering
    \setlength{\tabcolsep}{2pt}
    {%
    \begin{tabular}{@{}ccc@{}}
    \toprule
    \textbf{Box Search} & \textbf{CS} & \textbf{SS} \\ \midrule
    $L_\text{OD}$ & 4.65 & 12.56 \\
    $L_\text{box}$ & \textbf{6.74} & \textbf{13.58} \\ \bottomrule
    \end{tabular}%
    \caption{Performance in mAP(\%) with box searching by $L_\text{OD}$ only or $L_\text{box}$ (w/ $L_\text{OD}$, $L_\text{FVD}$, and $L_\text{BVC}$).}
    \label{tab:ablation_box_search}
    }
  \end{minipage}
\end{table}

\begin{table}[t]
\centering
\setlength{\tabcolsep}{2pt}
{%
\begin{tabular}{@{}c|cccccc@{}}
\toprule
\multirow{2}{*}{\textbf{Methods}} & \begin{tabular}[c]{@{}c@{}}\textbf{Target-}\\ \textbf{FT}\end{tabular} & 
\begin{tabular}[c]{@{}c@{}}\textbf{Proto-}\\ \textbf{Vote}\end{tabular}
& \begin{tabular}[c]{@{}c@{}}\textbf{PVAE-}\\ \textbf{Vote}\end{tabular}& \textbf{CP-Vote} & \textbf{GFS} & \textbf{Ours} \\ \midrule
Common  & 15.28    & 6.97       & 7.43       & 8.73      & 17.37     & \textbf{18.06}   \\
Novel   & 6.39     & 7.05       & 7.53       & 7.17      & 1.16      & \textbf{11.11}   \\
Overall & 10.20    & 7.02       & 7.49       & 7.84      & 8.10      & \textbf{14.09}   \\ \bottomrule
\end{tabular}
}
\caption{Performance in mAP(\%) of PV-RCNN for NuScenes $\rightarrow5$shot-KITTI. Please refer to the supplementary for specifics across other GCFS tasks.}
\label{tab:PVRCNN_NutoFsKi_ablation}
\end{table}

\begin{table}[t]
\centering
\setlength{\tabcolsep}{2pt}
{%
\begin{tabular}{@{}ccccccc|c@{}}
\toprule
\textbf{Shots} & \textbf{K=1} & \textbf{K=3} & \textbf{K=5} & \textbf{K=10} & \textbf{K=20} & \textbf{K=40} & \textit{Full-shot} \\ \midrule
Common  & 7.27  & 12.27 & 15.99 & 23.56 & 27.59 & 32.05 & 41.34 \\
Novel   & 0.57  & 7.76  & 11.72 & 12.21 & 17.49 & 21.55 & 18.35 \\
Overall & 3.44  & 9.70  & 13.55 & 17.08 & 21.82 & 26.05 & 28.21 \\ \bottomrule
\end{tabular}
}
\caption{Performances in mAP(\%) with different $K$. \textit{Full-shot} denotes the training on the complete KITTI \textit{train} set.}
\label{tab:K_shot_ablation}
\end{table}

\begin{table}[]
\centering
\setlength{\tabcolsep}{2pt}
\begin{tabular}{@{}c|ccc|cc@{}}
\toprule
\multirow{2}{*}{\textbf{Method}} & \multicolumn{3}{c|}{DA}  & \multicolumn{2}{c}{OVD} \\  
                        & SN    & ST3D++ & DenResamp & FnP        & \textbf{Ours-OVD}       \\ \midrule
\textbf{Common}                  & 12.09 & 21.00  & 14.89  & 10.59      & \textbf{22.25}      \\
\textbf{Novel}                   & -     & -      & -      & 2.66       & \textbf{8.26}       \\
\textbf{Overall}                 & -     & -      & -      & 6.06       & \textbf{14.26}      \\ \bottomrule
\end{tabular}
\caption{Comparison in mAP(\%) for OVD and DA methods in the unsupervised few-shot setting. }
\label{tab:OV_ablation}
\end{table}

\noindent \textbf{Component Ablation.} 
Table~\ref{tab:component_ablation} (a)$\rightarrow$(b) indicates that our adaptive prototype learning enhances performance in common and novel classes, illustrating its swift adaptation to limited samples in the target domain.
Applying our image-guided multi-modal fusion (a)$\rightarrow$(c) yields marked improvement, especially on novel classes, showing its boost on object recall. 
By combining both, our GCFS method achieves the highest performance, demonstrating the complementarity of the two approaches. 
Notably, removing MAML lowers AP to 12.35, and replacing our box search with FnP gives AP of 12.58, showing our method’s effectiveness.

We also conduct ablations on our proposed prototype learning and box searching components. Table~\ref{tab:ablation_CL_proto} shows that the contrastive loss boosts model performance, indicating its ability to swiftly adapt prototypes to few-shot data in the target domain. In Table~\ref{tab:ablation_box_search}, integrating $L_\text{OD}$ with $L_\text{FVD}$ and $L_\text{BVC}$ yields improvements on both CS and SS objects, showing $L_\text{FVD}$ and $L_\text{BVC}$ enhancing recall rates for objects with diverse structural complexities, thereby further optimizing model effectiveness.

\noindent \textbf{Ablation on Detection Backbone.} 
We further evaluate our GCFS framework using the point-voxel-hybrid detector PV-RCNN. As shown in Table~\ref{tab:PVRCNN_NutoFsKi_ablation}, our approach consistently outperforms others across all three metrics, demonstrating the generalizability of our solution.

\noindent \textbf{Ablation on Numbers of Shots.} 
Table~\ref{tab:K_shot_ablation} shows that our method scales well with increasing $K$.  
At $K=40$, the overall performance approaches the full-shot, narrowing the supervision gap. Despite a reasonable gap in common-class performance due to limited data, the novel-class performance surpasses the full-shot result, due to class imbalance in full-shot training and our image-guided design enhancing novel object discovery. These results confirm the scalability and generalization of our method under limited supervision.


\noindent \textbf{Unsupervised few-shot ablation with OVD and DA methods.} 
We establish an unsupervised few-shot setting with no annotations for all classes. Our approach is benchmarked against the SOTA 3D-OVD solution
and well-established 3D-DA methods in Table~\ref{tab:OV_ablation}.  
To create an OVD version of our method, we incorporate a physical-aware box searcher to generate high-quality pseudo-labels for target-specific training.
Compared to OVD and DA methods, our OVD method achieves the highest mAPs, 
showing strong domain gap bridging capability and high learning efficiency from unlabeled samples. Please refer to the supplementary for implementation and result details.

\section{Conclusion}
\label{sec:conclusion} 
This paper tackled the generalized cross-domain few-shot task in 3D object detection  
and introduced the first GCFS solution. 
Beyond achieving state-of-the-art performance on four GCFS benchmarks, our work demonstrated a generalizable approach to few-shot 3D adaptation, grounded in the idea that semantic alignment across modalities and domains could be achieved by combining 2D open-set priors with 3D structural cues and few-shot supervision. We believed this framework opens new possibilities for 3D perception systems that must continually adapt to new environments and emerging object types, without relying on exhaustive data collection or domain-specific engineering.

\section*{Acknowledgements}
This work was supported in part by JST CRONOS Grant (No. JPMJCS24K8), JSPS KAKENHI Grant (No.JP21H04877, No.JP23H03372, and No.JP24K02920), Canada CIFAR AI Chairs Program, the Natural Sciences and Engineering Research Council of Canada, and the Autoware Foundation.

\small
\bibliography{aaai2026}

\appendix
\clearpage
\setcounter{page}{1}

In this supplementary material, we provide additional details on our proposed image-guided multi-modal fusion module and the optimization-based meta-learning scheme in Section~\ref{sec:appendix_methodology}. In addition, we provide further details on the benchmark settings and method implementations, along with a comprehensive presentation of the experimental results in Section~\ref{sec:appendix_experiment}.

\section{Details on Methodology}
\label{sec:appendix_methodology}
\subsection{Image-guided Multi-modal Fusion}
\noindent \textbf{Box searching calculation.} 
According to the center $[x, y, z]$ and heading angle $\theta$ of each object, the corresponding rotation transformation matrix \textbf{T} is defined as: 
\begin{equation}
\textbf{T} = \begin{bmatrix}
\cos \theta & -\sin \theta & 0 & t_y \sin \theta - t_x \cos \theta  \\ \\
\sin \theta & \cos \theta & 0 & -t_x \sin \theta - t_y \cos \theta \\
0 & 0 & 1 & -z \\
\end{bmatrix}.
\end{equation}
Regarding the FVD loss, we utilize the LiDAR's view angle to obtain the front-reviewed boundary $\textbf{FB}$ for each class. Specifically, for given box $[x, y, z, \theta]$, we first get the yaw angle by $\alpha = \arctan(\frac{x}{y})$. Then, we obtain the view angle $\phi=\alpha-\theta$, which indicates the LiDAR scanning direction w.r.t. the search box. Regarding $\phi$, we define $\textbf{FB}$ via the prior box length $l$ and width $w$:

\begin{multline}
\text{\textbf{FB}} = \left[ \frac{l}{2} S_{l}, \frac{w}{2} S_{w} \right], \\ \text{s.t. } [S_{l}, S_{w}] =
\begin{cases}
    \left[-1, -1\right], & 0 < \phi \leq \frac{\pi}{2} \\
    \left[1, -1\right], & \frac{\pi}{2} < \phi \leq \pi \\
    \left[1, 1\right], & \pi < \phi < \frac{3\pi}{2} \\
    \left[-1, 1\right], & \frac{3\pi}{2} < \phi \leq 2\pi
\end{cases}
\end{multline}
\noindent which indicates the faces of the box front-viewed by scanning LiDARs.

\subsection{Optimization-based Meta-learning}
\begin{figure}[h]
  \centering
   \includegraphics[width=0.8\linewidth]{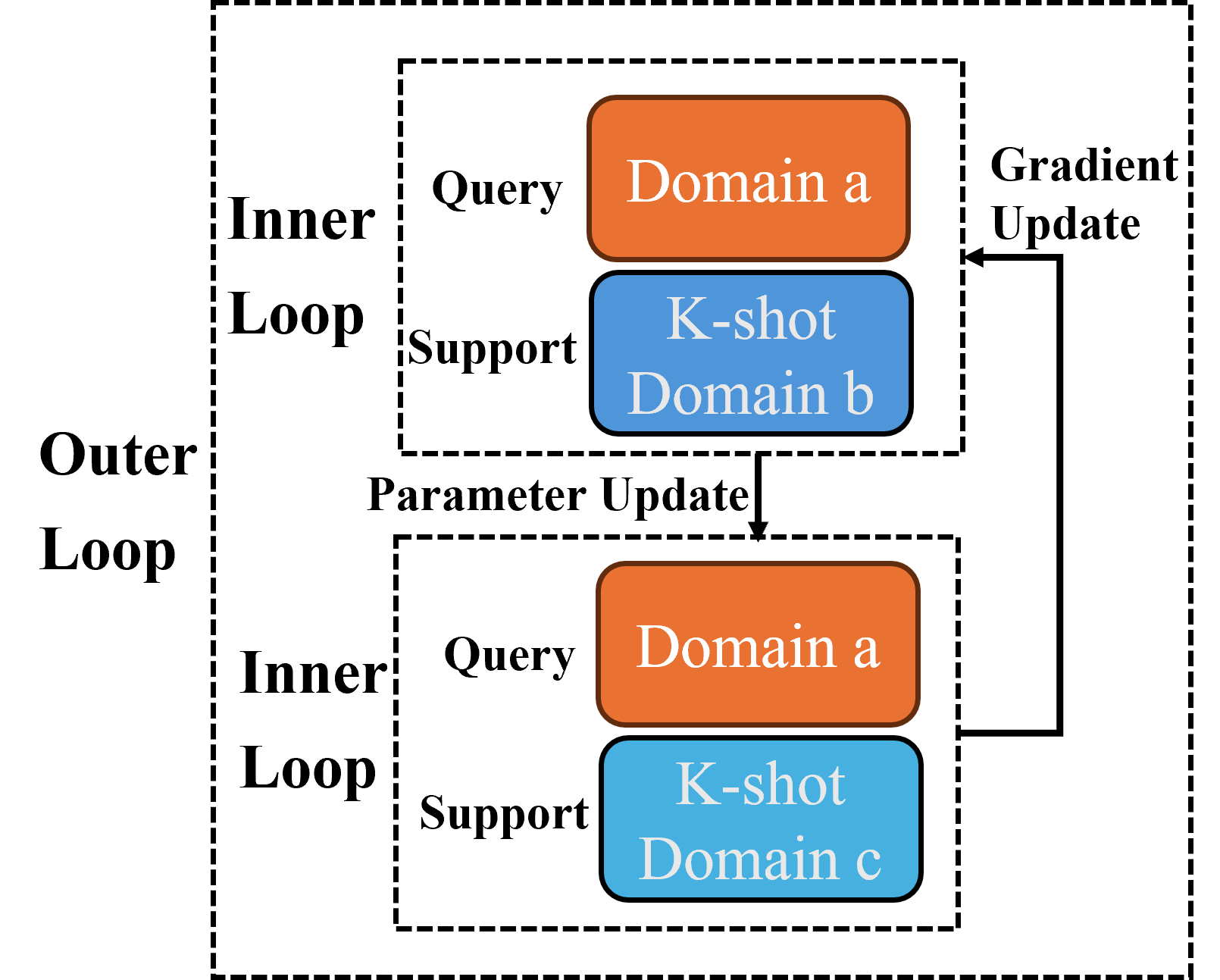}
   \caption{Meta-learning scheme simulating few-shot learning with domain gaps.}
   \label{fig: meta_learning}
\end{figure}

In meta-training, we utilize the sufficient data of common classes $\mathbb{C}_\text{com}$ and other classes $\mathbb{C}_\text{nov}^s$ to simulate the target few-shot fine-tuning on $\mathbb{C}_\text{com}$ and $\mathbb{C}_\text{nov}^t$. Specifically, we first randomly sample 
$N_\text{nov}$ classes from $\mathbb{C}_\text{nov}^s$, where $N_\text{nov}$ is the class number of $\mathbb{C}_\text{nov}^t$. Then, as shown in Figure~\ref{fig: meta_learning}, in the inner loop, we set up a $K$-shot cross-domain detection task, covering common classes and sampled $N_\text{nov}$ classes. For the meta-task, we utilize data augmentations on support data to simulate domain gaps regarding source-trained prototypes transferred to target data. Like MAML~\cite{finn2017model}, we design the outer loop, where we run one inner loop to get the one-step updated parameter, run another inner loop to get the two-step gradient, and use the two-step gradient to update the original model parameters. As in \cite{finn2017model}, this inner-outer-loop meta-learning will find a set of parameters and prototypes $\textbf{F}_\text{proto}$ with the quick adaptation to a new few-shot learning task (covering common classes and novel classes) in a new domain via limited target data.

\begin{table*}[]
\centering
\caption{Performance comparison in mAP(\%) on VoxelRCNN detector in the NuScenes $\rightarrow5$shot-KITTI GCFS task. 
The \textbf{bold} values represent the best performance. Subscript values in parentheses are standard deviations. 
The \textbf{\textit{ped}} is for Pedestrian, \textbf{\textit{trk}} for Truck, \textbf{\textit{ps}} for Person\_sitting, \textbf{\textit{cyc}} for Cyclist, and \textbf{\textit{trm}} for Tram.
}
\label{tab:app_vn_NtoK}
\resizebox{\textwidth}{!}{%
\begin{tabular}{@{}cc|ccc|c|cccc|c|c@{}}
\toprule
\textbf{Methods} & \textbf{Venues} & \textbf{car} & \textbf{ped} & \textbf{trk} & \cellcolor{lightgray}\textbf{common} & \textbf{van} & \textbf{ps} & \textbf{cyc} & \textbf{trm} & \cellcolor{lightgray}\textbf{novel} & \cellcolor{lightgray}\textbf{overall} \\ \midrule
Target-FT & - & 30.08\textsubscript{(2.75)} & 7.08\textsubscript{(1.86)} & 1.17\textsubscript{(0.94)} & \cellcolor{lightgray}12.77\textsubscript{(1.85)} & 14.24\textsubscript{(1.61)} & 0.87\textsubscript{(0.3)} & 4.59\textsubscript{(2.01)} & 2.24\textsubscript{(0.71)} & \cellcolor{lightgray}5.48\textsubscript{(1.16)} & \cellcolor{lightgray}8.61\textsubscript{(1.45)} \\
Proto-Vote & NIPS’22 & 17.93\textsubscript{(5.88)} & 4.49\textsubscript{(1.07)} & 0.26\textsubscript{(0.22)} & \cellcolor{lightgray}7.56\textsubscript{(2.39)} & 11.36\textsubscript{(3.19)} & 0.02\textsubscript{(0.02)} & 9.86\textsubscript{(2.14)} & 1.7\textsubscript{(0.8)} & \cellcolor{lightgray}5.74\textsubscript{(1.54)} & \cellcolor{lightgray}6.52\textsubscript{(1.9)} \\
PVAE-Vote & NIPS’24 & 18.76\textsubscript{(6.17)} & 4.06\textsubscript{(1.64)} & 1.23\textsubscript{(0.51)} & \cellcolor{lightgray}8.01\textsubscript{(2.77)} & 12.07\textsubscript{(3.67)} & 0.05\textsubscript{(0.04)} & 11.22\textsubscript{(3.17)} & 2.17\textsubscript{(2)} & \cellcolor{lightgray}6.38\textsubscript{(2.22)} & \cellcolor{lightgray}7.08\textsubscript{(2.46)} \\
CP-Vote & PRCV’24 & 24.7\textsubscript{(4.78)} & 6.34\textsubscript{(1.51)} & 1.02\textsubscript{(0.54)} & \cellcolor{lightgray}10.69\textsubscript{(2.28)} & 11.2\textsubscript{(3.16)} & 0.57\textsubscript{(0.21)} & 15.5\textsubscript{(2.52)} & 4.08\textsubscript{(1.59)} & \cellcolor{lightgray}7.84\textsubscript{(1.87)} & \cellcolor{lightgray}9.06\textsubscript{(2.04)} \\
GFS-Det & arXiv’23 & 17.1\textsubscript{(3.27)} & \textbf{19.86}\textsubscript{(3.67)} & 1.54\textsubscript{(0.13)} & \cellcolor{lightgray}12.83\textsubscript{(2.36)} & 2.42\textsubscript{(0.55)} & 0.05\textsubscript{(0.02)} & 2.15\textsubscript{(0.92)} & 0.1\textsubscript{(0.02)} & \cellcolor{lightgray}1.18\textsubscript{(0.38)} & \cellcolor{lightgray}6.17\textsubscript{(1.23)} \\ \midrule
\textbf{Ours} & - & \textbf{37.71}\textsubscript{(2.41)} & 7.16\textsubscript{(1.89)} & \textbf{3.11}\textsubscript{(0.56)} & \cellcolor{lightgray}\textbf{15.99}\textsubscript{(1.62)} & \textbf{22.29}\textsubscript{(2.43)} & \textbf{1.54}\textsubscript{(0.28)} & \textbf{18.26}\textsubscript{(1.56)} & \textbf{4.79}\textsubscript{(1.33)} & \cellcolor{lightgray}\textbf{11.72}\textsubscript{(1.4)} & \cellcolor{lightgray}\textbf{13.55}\textsubscript{(1.5)} \\ \bottomrule
\end{tabular}%
}
\end{table*}

\begin{table*}[]
\centering
\caption{Performance comparison in mAP(\%) on PVRCNN++ detector in the NuScenes $\rightarrow5$shot-KITTI GCFS task. 
The \textbf{bold} values represent the best performance. Subscript values in parentheses are standard deviations. 
The \textbf{\textit{ped}} is for Pedestrian, \textbf{\textit{trk}} for Truck, \textbf{\textit{ps}} for Person\_sitting, \textbf{\textit{cyc}} for Cyclist, and \textbf{\textit{trm}} for Tram.}
\label{tab:app_pn_NtoK}
\resizebox{\textwidth}{!}{%
\begin{tabular}{@{}cc|ccc|c|cccc|c|c@{}}
\toprule
\textbf{Methods} & \textbf{Venues} & \textbf{car} & \textbf{ped} & \textbf{trk} & \cellcolor{lightgray}\textbf{common} & \textbf{van} & \textbf{ps} & \textbf{cyc} & \textbf{trm} & \cellcolor{lightgray}\textbf{novel} & \cellcolor{lightgray}\textbf{overall} \\ \midrule
Target-FT & - & 38.79\textsubscript{(4.5)} & 6.13\textsubscript{(0.8)} & 0.92\textsubscript{(0.79)} & \cellcolor{lightgray}15.28\textsubscript{(2.03)} & 11.3\textsubscript{(2)} & \textbf{2.08}\textsubscript{(1.29)} & 11.16\textsubscript{(2.7)} & 1.05\textsubscript{(0.86)} & \cellcolor{lightgray}6.39\textsubscript{(1.71)} & \cellcolor{lightgray}10.2\textsubscript{(1.85)} \\
Proto-Vote & NIPS’22 & 15.83\textsubscript{(3.21)} & 3.09\textsubscript{(0.25)} & 2\textsubscript{(0.82)} & \cellcolor{lightgray}6.97\textsubscript{(1.43)} & 13.26\textsubscript{(1.9)} & 0.61\textsubscript{(0.4)} & 12.55\textsubscript{(5.09)} & 1.78\textsubscript{(0.66)} & \cellcolor{lightgray}7.05\textsubscript{(2.02)} & \cellcolor{lightgray}7.02\textsubscript{(1.76)} \\
PVAE-Vote & NIPS’24 & 17.1\textsubscript{(3.22)} & 3.17\textsubscript{(0.99)} & 2.02\textsubscript{(0.95)} & \cellcolor{lightgray}7.43\textsubscript{(1.72)} & 14.12\textsubscript{(2.18)} & 0.28\textsubscript{(0.16)} & 13.67\textsubscript{(5.48)} & 2.05\textsubscript{(0.81)} & \cellcolor{lightgray}7.53\textsubscript{(2.16)} & \cellcolor{lightgray}7.49\textsubscript{(1.97)} \\
CP-Vote & PRCV’24 & 22.92\textsubscript{(4.26)} & 2.74\textsubscript{(1.26)} & 0.53\textsubscript{(0.41)} & \cellcolor{lightgray}8.73\textsubscript{(1.97)} & 12.55\textsubscript{(2.37)} & 0.55\textsubscript{(0.31)} & 13.31\textsubscript{(3.96)} & 2.29\textsubscript{(0.56)} & \cellcolor{lightgray}7.17\textsubscript{(1.8)} & \cellcolor{lightgray}7.84\textsubscript{(1.87)} \\
GFS-Det & arXiv’23 & 21.04\textsubscript{(2.2)} & \textbf{30.16}\textsubscript{(3.77)} & 0.9\textsubscript{(0.28)} & \cellcolor{lightgray}17.37\textsubscript{(2.08)} & 3.41\textsubscript{(0.31)} & 0.05\textsubscript{(0.01)} & 1.03\textsubscript{(0.21)} & 0.13\textsubscript{(0.02)} & \cellcolor{lightgray}1.16\textsubscript{(0.14)} & \cellcolor{lightgray}8.1\textsubscript{(0.97)} \\ \midrule
\textbf{Ours} & - & \textbf{44.92}\textsubscript{(2.61)} & 6.13\textsubscript{(0.67)} & \textbf{3.14}\textsubscript{(1.09)} & \cellcolor{lightgray}\textbf{18.06}\textsubscript{(1.46)} & \textbf{21.16}\textsubscript{(1.35)} & 0.86\textsubscript{(0.11)} & \textbf{18.05}\textsubscript{(2.03)} & \textbf{4.36}\textsubscript{(0.61)} & \cellcolor{lightgray}\textbf{11.11}\textsubscript{(1.03)} & \cellcolor{lightgray}\textbf{14.09}\textsubscript{(1.21)} \\ \bottomrule
\end{tabular}%
}
\end{table*}

\begin{table*}[]
\centering
\caption{Performance comparison in mAP(\%) on VoxelRCNN detector in the Waymo $\rightarrow5$shot-KITTI GCFS task. 
The \textbf{bold} values represent the best performance. Subscript values in parentheses are standard deviations. 
The \textbf{\textit{ped}} is for Pedestrian, \textbf{\textit{trk}} for Truck, \textbf{\textit{ps}} for Person\_sitting, \textbf{\textit{cyc}} for Cyclist, and \textbf{\textit{trm}} for Tram.}
\label{tab:app_vn_WtoK}
\resizebox{\textwidth}{!}{%
\begin{tabular}{@{}cc|ccc|c|cccc|c|c@{}}
\toprule
\textbf{Methods} & \textbf{Venues} & \textbf{car} & \textbf{ped} & \textbf{trk} & \cellcolor{lightgray} \textbf{common} & \textbf{van} & \textbf{ps} & \textbf{cyc} & \textbf{trm} & \cellcolor{lightgray} \textbf{novel} & \cellcolor{lightgray} \textbf{overall} \\ \midrule
Target-FT & - & 55.86\textsubscript{(2.93)} & 11.79\textsubscript{(1.08)} & 1.52\textsubscript{(0.89)} & \cellcolor{lightgray} 23.06\textsubscript{(1.64)} & 21.03\textsubscript{(3.49)} & \textbf{2.71}\textsubscript{(0.83)} & 25.54\textsubscript{(3.04)} & 0.59\textsubscript{(0.32)} & \cellcolor{lightgray} 12.47\textsubscript{(1.92)} & \cellcolor{lightgray} 17.01\textsubscript{(1.8)} \\
Proto-Vote & NIPS’22 & 39\textsubscript{(2.07)} & 9.45\textsubscript{(4.09)} & 3.64\textsubscript{(2.01)} & \cellcolor{lightgray} 17.36\textsubscript{(2.72)} & 20.24\textsubscript{(2.77)} & 1.26\textsubscript{(0.52)} & 23.79\textsubscript{(2.33)} & 3.02\textsubscript{(1.51)} & \cellcolor{lightgray} 12.08\textsubscript{(1.78)} & \cellcolor{lightgray} 14.34\textsubscript{(2.19)} \\
PVAE-Vote & NIPS’24 & 41.99\textsubscript{(2.91)} & 9.55\textsubscript{(4.12)} & 3.01\textsubscript{(1.79)} & \cellcolor{lightgray} 18.19\textsubscript{(2.94)} & 22.51\textsubscript{(4.17)} & 1.01\textsubscript{(0.48)} & 24.22\textsubscript{(2.51)} & 3.41\textsubscript{(1.64)} & \cellcolor{lightgray} 12.79\textsubscript{(2.2)} & \cellcolor{lightgray} 15.1\textsubscript{(2.52)} \\
CP-Vote & PRCV’24 & 41.35\textsubscript{(2.5)} & 7.67\textsubscript{(1.78)} & \textbf{3.95}\textsubscript{(2.92)} & \cellcolor{lightgray} 17.66\textsubscript{(2.4)} & 20.42\textsubscript{(2.31)} & 1.41\textsubscript{(0.45)} & 22.63\textsubscript{(3.09)} & 4.22\textsubscript{(1.59)} & \cellcolor{lightgray} 12.17\textsubscript{(1.86)} & \cellcolor{lightgray} 14.52\textsubscript{(2.09)} \\
GFS-Det & arXiv’23 & 21.96\textsubscript{(2.99)} & \textbf{42.83}\textsubscript{(4.37)} & 3.44\textsubscript{(0.98)} & \cellcolor{lightgray} 22.74\textsubscript{(2.78)} & 2.13\textsubscript{(0.79)} & 1.04\textsubscript{(0.3)} & 1.77\textsubscript{(0.39)} & 0.11\textsubscript{(0.04)} & \cellcolor{lightgray} 1.26\textsubscript{(0.38)} & \cellcolor{lightgray} 10.47\textsubscript{(1.41)} \\ \midrule
\textbf{Ours} & - & \textbf{57.36}\textsubscript{(1.78)} & 15.12\textsubscript{(3.04)} & 3.73\textsubscript{(1.26)} & \cellcolor{lightgray} \textbf{25.4}\textsubscript{(2.02)} & \textbf{28.7}\textsubscript{(1.69)} & 1.47\textsubscript{(0.18)} & \textbf{30.48}\textsubscript{(2.32)} & \textbf{10.36}\textsubscript{(2.46)} & \cellcolor{lightgray} \textbf{17.75}\textsubscript{(1.66)} & \cellcolor{lightgray} \textbf{21.03}\textsubscript{(1.82)}\\ \bottomrule
\end{tabular}%
}
\end{table*}

\begin{table*}[]
\centering
\caption{Performance comparison in mAP(\%) on PVRCNN++ detector in the Waymo $\rightarrow5$shot-KITTI GCFS task. 
The \textbf{bold} values represent the best performance. Subscript values in parentheses are standard deviations. 
The \textbf{\textit{ped}} is for Pedestrian, \textbf{\textit{trk}} for Truck, \textbf{\textit{ps}} for Person\_sitting, \textbf{\textit{cyc}} for Cyclist, and \textbf{\textit{trm}} for Tram.}
\label{tab:app_pn_WtoK}
\resizebox{\textwidth}{!}{%
\begin{tabular}{@{}cc|ccc|c|cccc|c|c@{}}
\toprule
\textbf{Methods} & \textbf{Venues} & \textbf{car} & \textbf{ped} & \textbf{trk} & \cellcolor{lightgray} \textbf{common} & \textbf{van} & \textbf{ps} & \textbf{cyc} & \textbf{trm} & \cellcolor{lightgray} \textbf{novel} & \cellcolor{lightgray} \textbf{overall} \\ \midrule
Target-FT & - & 59.18\textsubscript{(2.88)} & 12.12\textsubscript{(0.76)} & 1.74\textsubscript{(0.55)} & \cellcolor{lightgray} 24.35\textsubscript{(1.4)} & 24.94\textsubscript{(2.73)} & \textbf{2.25}\textsubscript{(0.75)} & 26.88\textsubscript{(2.57)} & 1.08\textsubscript{(0.86)} & \cellcolor{lightgray} 13.79\textsubscript{(1.73)} & \cellcolor{lightgray} 18.31\textsubscript{(1.59)} \\
Proto-Vote & NIPS’22 & 22.36\textsubscript{(3.93)} & 4.24\textsubscript{(0.71)} & 3.45\textsubscript{(1.84)} & \cellcolor{lightgray} 10.02\textsubscript{(2.16)} & 22.12\textsubscript{(2.03)} & 0.83\textsubscript{(0.43)} & 23.76\textsubscript{(2.99)} & 2.39\textsubscript{(0.52)} & \cellcolor{lightgray} 12.27\textsubscript{(1.49)} & \cellcolor{lightgray} 11.31\textsubscript{(1.78)} \\
PVAE-Vote & NIPS’24 & 25.42\textsubscript{(4.51)} & 5\textsubscript{(0.79)} & 3.67\textsubscript{(1.89)} & \cellcolor{lightgray} 11.36\textsubscript{(2.4)} & 23.05\textsubscript{(2.67)} & 1.26\textsubscript{(0.72)} & 23.92\textsubscript{(3.52)} & 3.13\textsubscript{(0.82)} & \cellcolor{lightgray} 12.84\textsubscript{(1.93)} & \cellcolor{lightgray} 12.21\textsubscript{(2.13)} \\
CP-Vote & PRCV’24 & 38.09\textsubscript{(6.66)} & 3.15\textsubscript{(1.04)} & 1.31\textsubscript{(0.68)} & \cellcolor{lightgray} 14.18\textsubscript{(2.79)} & 22.3\textsubscript{(3.93)} & 1.2\textsubscript{(0.29)} & 25.94\textsubscript{(6.01)} & 3.83\textsubscript{(0.71)} & \cellcolor{lightgray} 13.32\textsubscript{(2.74)} & \cellcolor{lightgray} 13.69\textsubscript{(2.76)} \\
GFS-Det & arXiv’23 & 26.09\textsubscript{(1.58)} & \textbf{43.02}\textsubscript{(2.55)} & 5.25\textsubscript{(1.25)} & \cellcolor{lightgray} 24.79\textsubscript{(1.79)} & 1.93\textsubscript{(0.33)} & 1.23\textsubscript{(0.3)} & 1.99\textsubscript{(0.23)} & 0.16\textsubscript{(0.05)} & \cellcolor{lightgray} 1.33\textsubscript{(0.23)} & \cellcolor{lightgray} 11.38\textsubscript{(0.9)} \\ \midrule
\textbf{Ours} & - & \textbf{59.2}\textsubscript{(1.8)} & 15.17\textsubscript{(1.56)} & \textbf{7.68}\textsubscript{(1.3)} & \cellcolor{lightgray} \textbf{27.35}\textsubscript{(1.55)} & \textbf{28.06}\textsubscript{(3.02)} & 1.7\textsubscript{(0.44)} & \textbf{27.01}\textsubscript{(2.79)} & \textbf{7.3}\textsubscript{(1.46)} & \cellcolor{lightgray} \textbf{16.02}\textsubscript{(1.93)} & \cellcolor{lightgray} \textbf{20.88}\textsubscript{(1.77)} \\ \bottomrule
\end{tabular}%
}
\end{table*}

\begin{table*}[]
\centering
\caption{Performance comparison in mAP(\%) on VoxelRCNN detector in the KITTI $\rightarrow5$shot-A2D2 GCFS task. 
The \textbf{bold} values represent the best performance. Subscript values in parentheses are standard deviations.
The \textbf{\textit{trk}} for Truck, \textbf{\textit{bcy}} is for Bicycle, and \textbf{\textit{uvc}} for Utility\_vehicle.
}
\label{tab:app_vn_KtoA}
\resizebox{\textwidth}{!}{%
\begin{tabular}{@{}cc|ccc|c|ccc|c|c@{}}
\toprule
\textbf{Methods} & \textbf{Venues} & \textbf{car} & \textbf{ped} & \textbf{trk} & \cellcolor{lightgray} \textbf{common} & \textbf{bcy} & \textbf{uvc} & \textbf{bus} & \cellcolor{lightgray} \textbf{novel} & \cellcolor{lightgray} \textbf{overall} \\ \midrule
Target-FT & - & 2.82\textsubscript{(0.43)} & 3.58\textsubscript{(1.06)} & 8.89\textsubscript{(1.8)} & \cellcolor{lightgray} 5.09\textsubscript{(1.1)} & 0.25\textsubscript{(0.08)} & 0.04\textsubscript{(0.03)} & 1.83\textsubscript{(0.4)} & \cellcolor{lightgray} 0.7\textsubscript{(0.17)} & \cellcolor{lightgray} 2.9\textsubscript{(0.64)} \\
Proto-Vote & NIPS’22 & 1.89\textsubscript{(0.27)} & 3.38\textsubscript{(0.69)} & 5.58\textsubscript{(1.49)} & \cellcolor{lightgray} 3.61\textsubscript{(0.81)} & 2.66\textsubscript{(0.72)} & 0.19\textsubscript{(0.05)} & 2.74\textsubscript{(0.68)} & \cellcolor{lightgray} 1.86\textsubscript{(0.48)} & \cellcolor{lightgray} 2.74\textsubscript{(0.65)} \\
PVAE-Vote & NIPS’24 & 1.8\textsubscript{(0.41)} & 3.07\textsubscript{(0.71)} & 5.42\textsubscript{(1.45)} & \cellcolor{lightgray} 3.43\textsubscript{(0.85)} & 2.67\textsubscript{(0.76)} & 0.1\textsubscript{(0.03)} & 3.14\textsubscript{(0.83)} & \cellcolor{lightgray} 1.97\textsubscript{(0.54)} & \cellcolor{lightgray} 2.7\textsubscript{(0.7)} \\
CP-Vote & PRCV’24 & 3.83\textsubscript{(1.06)} & 3.02\textsubscript{(0.66)} & 6.01\textsubscript{(1.05)} & \cellcolor{lightgray} 4.28\textsubscript{(0.92)} & \textbf{2.84}\textsubscript{(1.25)} & 0.38\textsubscript{(0.15)} & 4.94\textsubscript{(1.24)} & \cellcolor{lightgray} 2.72\textsubscript{(0.88)} & \cellcolor{lightgray} 3.5\textsubscript{(0.9)} \\
GFS & arXiv’23 & \textbf{5.13}\textsubscript{(0.49)} & \textbf{6.13}\textsubscript{(0.9)} & 1.91\textsubscript{(0.32)} & \cellcolor{lightgray} 4.39\textsubscript{(0.57)} & 0.15\textsubscript{(0.05)} & 0.02\textsubscript{(0.01)} & 0.48\textsubscript{(0.1)} & \cellcolor{lightgray} 0.22\textsubscript{(0.05)} & \cellcolor{lightgray} 2.3\textsubscript{(0.31)} \\ \midrule
\textbf{Ours} & - & 3.08\textsubscript{(0.49)} & 3.89\textsubscript{(0.49)} & \textbf{16.38}\textsubscript{(1.06)} & \cellcolor{lightgray} \textbf{7.78}\textsubscript{(0.68)} & 1.76\textsubscript{(0.22)} & \textbf{2.03}\textsubscript{(0.64)} & \textbf{11.89}\textsubscript{(0.96)} & \cellcolor{lightgray} \textbf{5.22}\textsubscript{(0.61)} & \cellcolor{lightgray} \textbf{6.5}\textsubscript{(0.64)} \\ \bottomrule
\end{tabular}%
}
\end{table*}

\begin{table*}[]
\centering
\caption{Performance comparison in mAP(\%) on PVRCNN++ detector in the KITTI $\rightarrow5$shot-A2D2 GCFS task. 
The \textbf{bold} values represent the best performance. Subscript values in parentheses are standard deviations.
The \textbf{\textit{trk}} for Truck, \textbf{\textit{bcy}} is for Bicycle, and \textbf{\textit{uvc}} for Utility\_vehicle.}
\label{tab:app_pn_KtoA}
\resizebox{\textwidth}{!}{%
\begin{tabular}{@{}cc|ccc|c|ccc|c|c@{}}
\toprule
\textbf{Methods} & \textbf{Venues} & \textbf{car} & \textbf{ped} & \textbf{trk} & \cellcolor{lightgray} \textbf{common} & \textbf{bcy} & \textbf{uvc} & \textbf{bus} & \cellcolor{lightgray} \textbf{novel} & \cellcolor{lightgray} \textbf{overall} \\ \midrule
Target-FT & - & 3.53\textsubscript{(0.91)} & 2.49\textsubscript{(0.58)} & 4.9\textsubscript{(1.52)} & \cellcolor{lightgray} 3.64\textsubscript{(1)} & 0.31\textsubscript{(0.07)} & 0.07\textsubscript{(0.04)} & 0.41\textsubscript{(0.18)} & \cellcolor{lightgray} 0.26\textsubscript{(0.09)} & \cellcolor{lightgray} 1.95\textsubscript{(0.55)} \\
Proto-Vote & NIPS’22 & 1.93\textsubscript{(0.43)} & 1.93\textsubscript{(0.39)} & 7.9\textsubscript{(1.6)} & \cellcolor{lightgray} 3.92\textsubscript{(0.81)} & 1.78\textsubscript{(0.29)} & 0.08\textsubscript{(0.04)} & 2.57\textsubscript{(1.16)} & \cellcolor{lightgray} 1.48\textsubscript{(0.5)} & \cellcolor{lightgray} 2.7\textsubscript{(0.65)} \\
PVAE-Vote & NIPS’24 & 1.71\textsubscript{(0.83)} & 2.05\textsubscript{(0.54)} & 8.15\textsubscript{(1.5)} & \cellcolor{lightgray} 3.97\textsubscript{(0.96)} & 1.88\textsubscript{(0.51)} & 0.11\textsubscript{(0.08)} & 2.04\textsubscript{(1.02)} & \cellcolor{lightgray} 1.34\textsubscript{(0.54)} & \cellcolor{lightgray} 2.66\textsubscript{(0.75)} \\
CP-Vote & PRCV’24 & 2.69\textsubscript{(0.77)} & 3.14\textsubscript{(1.32)} & 6.77\textsubscript{(1.21)} & \cellcolor{lightgray} 4.2\textsubscript{(1.1)} & 2.49\textsubscript{(1.08)} & 0.06\textsubscript{(0.04)} & 3.87\textsubscript{(0.78)} & \cellcolor{lightgray} 2.14\textsubscript{(0.63)} & \cellcolor{lightgray} 3.17\textsubscript{(0.87)} \\
GFS & arXiv’23 & 3.74\textsubscript{(0.7)} & \textbf{4.2}\textsubscript{(0.59)} & 2.42\textsubscript{(0.39)} & \cellcolor{lightgray} 3.46\textsubscript{(0.56)} & 0.24\textsubscript{(0.04)} & 0.06\textsubscript{(0.03)} & 0.13\textsubscript{(0.05)} & \cellcolor{lightgray} 0.14\textsubscript{(0.04)} & \cellcolor{lightgray} 1.8\textsubscript{(0.3)} \\ \midrule
\textbf{Ours} & - & \textbf{3.8}\textsubscript{(1.02)} & 3.52\textsubscript{(0.46)} & \textbf{21.94}\textsubscript{(1.78)} & \cellcolor{lightgray} \textbf{9.75}\textsubscript{(1.09)} & \textbf{2.88}\textsubscript{(0.35)} & \textbf{3.14}\textsubscript{(0.85)} & \textbf{11.28}\textsubscript{(1.19)} & \cellcolor{lightgray} \textbf{5.76}\textsubscript{(0.8)} & \cellcolor{lightgray} \textbf{7.76}\textsubscript{(0.94)} \\ \bottomrule
\end{tabular}%
}
\end{table*}

\begin{table*}[]
\centering
\caption{Performance comparison in mAP(\%) on VoxelRCNN detector in the KITTI $\rightarrow5$shot-Argo2 GCFS task. 
The \textbf{bold} values represent the best performance. Subscript values in parentheses are standard deviations.
The \textbf{\textit{mtc}} for Motorcycle, \textbf{\textit{tcn}} is for Traffic\_cone, \textbf{\textit{lve}} for Large\_vehicle, and \textbf{\textit{cbl}} for Construction\_barrel.}
\label{tab:app_vr_KtoArgo2}
\resizebox{\textwidth}{!}{%
\begin{tabular}{cc|ccc|c|ccccccc|c|c}
\toprule
\textbf{Methods} & \textbf{Venues} & \textbf{car}    & \textbf{ped}   & \textbf{trk}   & \textbf{common} & \textbf{bcy}   & \textbf{bus}   & \textbf{mtc}   & \textbf{tcn}   & \textbf{lve}   & \textbf{cbl}   & \textbf{sign}  & \textbf{novel} & \textbf{overall} \\ \midrule
Target-FT        & -               & 4.70\textsubscript{(0.32)}   & 1.66\textsubscript{(0.12)} & 3.20\textsubscript{(0.12)}  & \cellcolor{lightgray}3.18\textsubscript{(0.19)}  & 0.10\textsubscript{(0.07)}  & 2.96\textsubscript{(0.51)} & 0.01\textsubscript{(0.01)} & 0.03\textsubscript{(0.01)} & 0.14\textsubscript{(0.10)} & 0.12\textsubscript{(0.09)} & 0.99\textsubscript{(0.09)} & \cellcolor{lightgray}0.62\textsubscript{(0.12)} & \cellcolor{lightgray}1.39\textsubscript{(0.14)}   \\
Proto-Vote       & NIPS’22         & 4.16\textsubscript{(1.45)} & 1.40\textsubscript{(0.55)}  & 4.42\textsubscript{(0.68)} & \cellcolor{lightgray}3.33\textsubscript{(0.90)}   & \textbf{1.98}\textsubscript{(0.87)} & 2.19\textsubscript{(0.34)} & 0.49\textsubscript{(0.04)} & 0.14\textsubscript{(0.10)}  & 0.38\textsubscript{(0.39)} & 0.57\textsubscript{(0.44)} & 0.53\textsubscript{(0.36)} & \cellcolor{lightgray}0.90\textsubscript{(0.36)}  & \cellcolor{lightgray}1.63\textsubscript{(0.52)}   \\
PVAE-Vote        & NIPS’24         & 4.64\textsubscript{(1.92)} & 1.54\textsubscript{(0.57)} & 3.11\textsubscript{(0.61)} & \cellcolor{lightgray}3.10\textsubscript{(1.03)}   & 1.21\textsubscript{(0.14)} & 3.45\textsubscript{(0.43)} & 0.23\textsubscript{(0.04)} & 0.38\textsubscript{(0.08)} & 0.63\textsubscript{(0.53)} & 0.42\textsubscript{(0.22)} & 0.14\textsubscript{(0.39)} & \cellcolor{lightgray}0.92\textsubscript{(0.26)} & \cellcolor{lightgray}1.58\textsubscript{(0.49)}   \\
CP-Vote          & PRCV’24         & 3.92\textsubscript{(1.52)} & 1.26\textsubscript{(0.60)}  & 3.00\textsubscript{(0.48)} & \cellcolor{lightgray}2.72\textsubscript{(0.87)}  & 1.15\textsubscript{(0.58)} & 3.51\textsubscript{(0.45)} & 0.16\textsubscript{(0.05)} & 0.24\textsubscript{(0.26)} & 0.27\textsubscript{(0.18)} & 0.61\textsubscript{(0.59)} & 0.55\textsubscript{(0.41)} & \cellcolor{lightgray}0.93\textsubscript{(0.36)} & \cellcolor{lightgray}1.47\textsubscript{(0.51)}   \\
GFS              & arXiv’23        & \textbf{10.57}\textsubscript{(0.11)} & \textbf{4.82}\textsubscript{(0.09)} & 2.96\textsubscript{(0.01)} & \cellcolor{lightgray}6.11\textsubscript{(0.07)}  & 0.07\textsubscript{(0.02)} & 0.02\textsubscript{(0)}    & 0.04\textsubscript{(0.01)} & 0.08\textsubscript{(0.03)} & 0.02\textsubscript{(0.00)} & 0.00\textsubscript{(0.00)}       & 0.00\textsubscript{(0.00)}       & \cellcolor{lightgray}0.03\textsubscript{(0.01)} & \cellcolor{lightgray}1.86\textsubscript{(0.03)}   \\ \midrule
\textbf{Ours}    & -               & 7.78\textsubscript{(0.22)} & 4.38\textsubscript{(0.12)} & \textbf{7.97}\textsubscript{(0.13)} & \cellcolor{lightgray}\textbf{6.71}\textsubscript{(0.16)}  & 1.38\textsubscript{(0.05)} & \textbf{8.25}\textsubscript{(0.98)} & \textbf{0.63}\textsubscript{(0.01)} & \textbf{0.58}\textsubscript{(0.03)} & \textbf{0.97}\textsubscript{(0.11)} & \textbf{0.70}\textsubscript{(0.04)}  & \textbf{1.97}\textsubscript{(0.01)} & \cellcolor{lightgray}\textbf{2.07}\textsubscript{(0.17)} & \cellcolor{lightgray}\textbf{3.46}\textsubscript{(0.17)}   \\ \bottomrule
\end{tabular}%
}
\end{table*}

\section{Details on Experiments}
\label{sec:appendix_experiment}
\subsection{Experiment Settings} 
\noindent\textbf{GCFS benchmark settings.}
Nuscenes~\cite{caesar2020nuscenes} contains labeled point cloud samples collected by a 32-beam LiDAR, mainly covering classes: \textit{Car, Pedestrian, Truck, Bicycle, Barrier, Construction\_vehicle, Bus, Trailer, Motorcycle, Traffic\_cone, etc.} We use the \textit{train} set ($\sim$28K samples) including 3 common classes for model pertaining and the \textit{train} set including 3 common classes and the rest 7 classes for meta pertaining.
Waymo~\cite{sun2020scalability} utilizes a 64-beam spinning-scanning LiDAR and 4 forward-scanning LiDARs to collect point cloud samples. The original Waymo contains 4 classes, i.e. \{\textit{Vehicle, Pedestrian, Cyclist, Sign}\}, where ``Vehicle'' covers cars, trucks, buses, motorcycles, bicycles, etc. To align the class categories in cross-domain settings, we refine the ``Vehicle'' class labels into 5 different classes \{\textit{Car, Truck, Bus, Motorcycle, Bicycle}\}. Specifically, with the help of finer-grained segmentation annotations (involving those 5 vehicle classes), we use the point-wise segmentation label to relabel vehicle objects into the object category with the highest number of points. For refined Waymo, we use the segmentation-involved $\sim$24K \textit{train} samples, including 3 common classes for model pertaining and \textit{train} samples including 3 common classes and 4 source-specific classes \{\textit{Bus, Motorcycle, Bicycle, Sign}\} for meta pertaining.
KITTI~\cite{geiger2012we} contains $\sim$7K labeled samples collected by a 64-beam LiDAR, covering classes: \textit{Car, Pedestrian, Truck, Van, Person\_sitting, Cyclist, Tram}. 
A2D2~\cite{geyer2020a2d2} contains $\sim$12K labeled samples collected by a 16-beam LiDAR, mainly covering classes: \textit{Car, Pedestrian, Truck, Bicycle, Utility\_vehicle, Bus, etc.} 
Argoverse 2~\cite{wilson2023argoverse} provides 1000 scenes and $\sim$134K labeled LiDAR frames collected by the top-mounted 32-beam spinning LiDAR, covering a wide variety of urban scenes and diverse object categories such as \textit{Vehicle, Pedestrian, Bus, Bicycle, Motorcycle, Construction\_barrel, and Construction\_vehicle}.

For FS-KITTI, we randomly select samples from the \textit{train} set to form the training data, including $K$-shot objects for each class, and use the complete \textit{val} set for model evaluation. For FS-A2D2, we first randomly select 6 out of 12 sequences and randomly select samples to form the training data, including $K$-shot objects for each class. We take the other 6 sequences as \textit{test} data for model evaluation. Since the samples of FS-A2D2 are temporarily sequential, we uniformly sample 50\% of \textit{test} data for computational efficiency during model evaluation. Also, we remove a small number of test samples with erroneous annotations (e.g., labeling ``Car'' as ``Pedestrian'') and ambiguous class labels (e.g., labeling ``Bus'' as ``Truck'') for a fair evaluation. Code implementation and more detailed information on the data set split are available in our codebase. 

Regarding the evaluation metrics for all methods, the IoU thresholds for common classes are [\textit{Car:0.7, Pedestrian:0.5, Truck:0.5}].
For novel object categories, regarding differences in structure, size, and semantic ambiguity, prior works in 3D object detection have adopted different IoU thresholds for different objects, such as 0.5~\cite{tang2024prototypical,yang2022st3d++}, 0.3~\cite{baur2024liso,gambashidze2024weak}, and 0.25~\cite{tang2024prototypical,zhao2022prototypical}. In our work, we follow this principle and use 0.5 and 0.3 based on object difficulty.
In FS-KITTI, for novel classes with regular structure and size [\textit{Van, Cyclist, Tram}], we use 0.5, while for the structurally complex and semantically confusing Person\_sitting, we apply 0.3.
In the more challenging FS-A2D2 task, which features 16-beam fixed LiDAR, we adopt a uniform IoU = 0.3 for all novel classes: [\textit{Bicycle, Utility\_vehicle, Bus}] (no overlap with FS-KITTI novel classes).
For FS-Argo2, we reuse the same IoU thresholds as in FS-KITTI and FS-A2D2 for shared novel classes. For the remaining novel categories, we use 0.5 for well-defined objects [\textit{Construction\_barrel, Traffic\_cone, Large\_vehicle, Motorcycle}] and 0.3 for small objects \textit{Sign}.
Regarding the confidence score threshold, we adopt 0.1 for FS-KITTI and FS-Argo2 tasks and 0.001 for the more challenging KITTI$\rightarrow$FS-A2D2 task. 
Note that, for FS-KITTI, we record the average AP across all difficulty levels (i.e., Easy, Moderate, and Hard), while FS-A2D2 and FS-Argo2 do not define difficulty levels, so we record standard AP regarding all objects. Besides the widely used AP and mAP, we also adopt 2D accuracy metrics to further explore the performance of our methods (see Table~\ref{tab: BEV_BBox_comp}).

\noindent\textbf{Implementation details.} For the consistency of input point clouds across datasets, we unify the LiDAR coordinate system of all datasets by setting the origin on the ground. We adopt the point cloud range of [$-75.2m,-75.2m,-2m,75.2m,75.2m,4m$] and the voxel size of [$0.1m,0.1m,0.15m$]. For \textit{ground-truth sampling augmentation}, we utilize its image-involved version implemented for KITTI in \cite{song2024robofusion} and extend it to A2D2. Regarding the class-specific attention module, the head number is 4, and the dropout rate is 0.1. During meta-training, we apply \textit{point density-resampling}~\cite{li2025domain} on support data to enlarge domain shifts between query and support data. All experiments are conducted on 2× GeForce RTX-3090 with a total memory of 48GB. Our code implementation is 
based on the codebase of OpenPCDet~\cite{openpcdet2020} and RoboFusion~\cite{song2024robofusion}. 
For full-shot target learning, we use the training setting the same as the pre-training setting (e.g., augmentation, learning rate, optimizer, hyper-parameters) as in Section~\ref{sec:experiment}, and the epoch numbers of the full-shot training epoch are 80 for KITTI and A2D2, and 6 for Argoverse 2.
For meta-training, epoch numbers are limited to 5 for NuScenes and Waymo, and 15 for KITTI, to obtain swiftly-adaptive model weights.   
Batch sizes are 2 during pre-training and meta-training and 1 in few-shot fine-tuning and testing. $\lambda$, and $\lambda_1$, $\lambda_2$ are set to 1.0, 0.2, and 0.2.  The temperature $\tau$ in the InfoNCE loss is set to 0.07.

\begin{table*}[t]
\centering
\caption{Component ablations in mAP(\%) for all classes. (\textbf{Image-Fusion} is our proposed image-guided multi-modal fusion and \textbf{CL-Proto} is our proposed contrastive-learning-enhanced prototype learning.)}
\label{tab:appendix_component_ablation}
\resizebox{0.85\textwidth}{!}{%
\begin{tabular}{@{}c|ccc|ccc|c|cccc|c|c@{}}
\toprule
 & \begin{tabular}[c]{@{}c@{}}\textbf{Target-}\\ \textbf{FT}\end{tabular} & \begin{tabular}[c]{@{}c@{}}\textbf{Image-}\\ \textbf{Fusion}\end{tabular} & \begin{tabular}[c]{@{}c@{}}\textbf{CL-}\\ \textbf{Proto}\end{tabular} & \textbf{car} & \textbf{ped} & \textbf{trk} & \cellcolor{lightgray} \textbf{common} & \textbf{van} & \textbf{ps} & \textbf{cyc} & \textbf{trm} & \cellcolor{lightgray} \textbf{novel} & \cellcolor{lightgray}\textbf{overall} \\ \midrule
(a) & \checkmark &  &  & 30.08 & 7.08 & 1.17 & \cellcolor{lightgray}12.77 & 14.24 & 0.87 & 4.59 & 2.24 & \cellcolor{lightgray} 5.48 & \cellcolor{lightgray} 8.61 \\
(b) & \checkmark &  & \checkmark & 37.07 & 6.04 & 1.29 & \cellcolor{lightgray} 14.80 & 14.52 & 0.95 & 14.40 & 2.52 & \cellcolor{lightgray} 8.10 & \cellcolor{lightgray} 10.97 \\
(c) & \checkmark & \checkmark &  & 35.40 & \textbf{7.21} & 1.47 & \cellcolor{lightgray} 14.69 & 21.16 & 1.45 & \textbf{19.00} & 3.08 & \cellcolor{lightgray} 11.17 & \cellcolor{lightgray} 12.68 \\ \midrule
(d) & \checkmark & \checkmark & \checkmark & \textbf{37.71} & 7.16 & \textbf{3.11} & \cellcolor{lightgray} \textbf{15.99} & \textbf{22.29} & \textbf{1.54} & 18.26 & \textbf{4.79} & \cellcolor{lightgray} \textbf{11.72} & \cellcolor{lightgray} \textbf{13.55} \\ \bottomrule
\end{tabular}%
}
\end{table*}

\begin{table*}[t]
\centering
\caption{Performances in mAP(\%) with different $K$ for all classes. (\textit{Full-shot} denotes the detection model trained on all KITTI \textit{train} data following pre-training settings.)}
\label{tab:appendix_K_shots}
\resizebox{0.68\textwidth}{!}{%
\begin{tabular}{@{}cccc|c|cccc|c|c@{}}
\toprule
\textbf{K} & \textbf{car} & \textbf{ped} & \textbf{trk} & \cellcolor{lightgray} \textbf{common} & \textbf{van} & \textbf{ps} & \textbf{cyc} & \textbf{trm} & \cellcolor{lightgray} \textbf{novel} & \cellcolor{lightgray} \textbf{overall} \\ \midrule 
1 & 19.14 & 1.62 & 1.04 & \cellcolor{lightgray}7.27 & 1.88 & 0.00 & 0.40 & 0.00 & \cellcolor{lightgray}0.57 & \cellcolor{lightgray}3.44 \\
3 & 27.91 & 6.28 & 2.62 & \cellcolor{lightgray}12.27 & 15.62 & 0.06 & 9.96 & 5.42 & \cellcolor{lightgray}7.76 & \cellcolor{lightgray}9.70 \\
5 & 37.71 & 7.16 & 3.11 & \cellcolor{lightgray}15.99 & 22.29 & 1.54 & 18.26 & 4.79 & \cellcolor{lightgray}11.72 & \cellcolor{lightgray}13.55 \\
10 & 55.64 & 9.38 & 5.67 & \cellcolor{lightgray}23.56 & 19.04 & 0.85 & 26.34 & 2.60 & \cellcolor{lightgray}12.21 & \cellcolor{lightgray}17.08 \\
20 & 56.20 & 19.38 & 7.19 & \cellcolor{lightgray}27.59 & 27.92 & 2.23 & 33.82 & 6.01 & \cellcolor{lightgray}17.49 & \cellcolor{lightgray}21.82 \\ 
40 & 62.68 & 20.40 & 13.08 & \cellcolor{lightgray}32.05 & 37.89 & 2.55 & 39.45 & 6.30 & \cellcolor{lightgray}21.55 & \cellcolor{lightgray}26.05 \\
\midrule
\textit{Full-shot} & 80.64 & 40.53 & 2.85 & \cellcolor{lightgray}41.34 & 41.79 & 0.22 & 30.68 & 0.72 & \cellcolor{lightgray}18.35 & \cellcolor{lightgray}28.21 \\ \bottomrule
\end{tabular}%
}
\end{table*}

\noindent \textbf{Compared methods.} Indoor FSL methods (i.e., Proto-Vote~\cite{zhao2022prototypical}, PVAE-Vote~\cite{tang2024prototypical}, and CP-Vote~\cite{li2024cp}) are mainly designed for the detection of novel classes. We extend their processing to common classes to fit the GCFS tasks. Since Proto-Vote~\cite{zhao2022prototypical} is implemented for the indoor RGBD-based data with VoteNet~\cite{qi2019deep} as the base detection model, we extend it to the outdoor LiDAR-based data with the VoxelRCNN~\cite{deng2021voxel} as the base detection model, following our experiment setting. Considering no public codebase for PVAE-Vote and CP-Vote, we follow the paper methodologies and implementation details in the papers and extend them to the GCFS tasks.
For PVAE-Vote, given that the instability of VAE training is particularly pronounced in outdoor sparse and various point clouds, we incorporate the skip-connection architecture similar to ResNet, which enables VAE branches to learn residuals, thereby enhancing the stability of few-shot training. Regarding outdoor GFSL method GFS-Det with no public codebase, we follow the paper methodology and implementation details in \cite{liu2023generalized} to extend it to the GCFS tasks. DenResamp ~\cite{li2025domain} proposes a single-domain generalization method that utilizes density-resampling-based augmentation and test-time adaptation to bridge density-related domain gaps. We explore its domain-adaptive version developed in the paper~\cite{li2025domain} as a 3D-DA method.

\noindent \textbf{Unsupervised few-shot experiment.} We establish an unsupervised few-shot setting extending from the supervised NuScenes$\rightarrow5$shot-KITTI GCFS task, to form an unsupervised GCFS task where no box annotations are available as ground-truth labels for all classes. 
Please note that in the supervised GCFS task, we sample objects from the point cloud, as well as 2D and 3D ground truth annotations, to ensure compliance with the 5-shot setting. For images in the FS-dataset, the sampled 2D ground truth annotations serve as the guidance for the strict K-shot object box retrieval. However, in the unsupervised GCFS task, 2D ground truth annotations are unavailable. Consequently, we relax the K-shot constraint in the unsupervised GCFS task by retaining all objects. In the unsupervised GCFS task, object numbers are \{\textit{Car}: 59, \textit{Van}: 15, \textit{Truck}: 6, \textit{Cyclist}: 14, \textit{Pedestrian}: 25, \textit{Person\_sitting}: 8, \textit{Tram}: 12\}.
Leveraging only the prior box size, we extend our method by incorporating our box searcher to generate high-quality pseudo-labels on target data. Then, we use pseudo-labels with target data to train the model for adapting to target common and novel classes. During target data training, we don't include our box-searching module to avoid the model overfitting the pseudo-labels searched by our box-searching module, and only include the box-searching module during model testing.
We benchmark our approach against two main categories of methods, the OVD approach and DA methods, to explore their performance under the few-shot constraint. As a SOTA OVD model, FnP~\cite{etchegaray2024find} also relies on prior box size for box searching. As in ~\cite{etchegaray2024find}, 3D pseudo-labels are acquired by the greedy box seeker and greedy box oracle module processing the 2D box candidates generated by GLIP~\cite{li2021grounded}. Then, the 3D pseudo-labels are propagated via a remote propagator for model fine-tuning on target data. We also include the well-established 3D-DA methods for comparison. As in \cite{wang2020train}, via the box size prior, SN is used as an augmentation on source data during the model pre-training. As in \cite{yang2022st3d++}, ST3D++ uses random object scaling on source data during model pre-training and hybrid quality-aware pseudo-label generation during model self-training with target unlabeled few-shot data, following our pre-training and fine-tuning settings, respectively. Via weak supervision by the target box prior, SN ~\cite{wang2020train} leverages box-size-related data
augmentation to de-bias the impact of different object sizes
on model generalization.

\begin{table*}[t]
\centering
\caption{Comparison in mAP(\%) with OVD and DA methods under the unsupervised few-shot setting for all classes. }
\label{tab:appendix_OV_experiment}
\resizebox{0.87\textwidth}{!}{%
\begin{tabular}{@{}ccc|ccc|c|cccc|c|c@{}}
\toprule
\multicolumn{2}{c}{\textbf{Method}} & \textbf{Venus} & \textbf{car} & \textbf{ped} & \textbf{trk} & \cellcolor{lightgray}\textbf{common} & \textbf{van} & \textbf{ps} & \textbf{cyc} & \textbf{trm} & \cellcolor{lightgray}\textbf{novel} & \cellcolor{lightgray}\textbf{overall} \\ \midrule
\multirow{2}{*}{DA} & SN & CVPR'20 & 19.96 & 15.11 & 1.20 & \cellcolor{lightgray}12.09 & - & - & - & - & \cellcolor{lightgray}- & \cellcolor{lightgray}- \\
& ST3D++ & PAMI'22 & \textbf{56.68} & 4.66 & 1.65 & \cellcolor{lightgray}21.00 & - & - & - & - & \cellcolor{lightgray}- & \cellcolor{lightgray}- \\ 
& DenResamp & ECCV'24 & 18.08 & \textbf{24.96} & 1.63 & \cellcolor{lightgray}14.89 & - & - & - & - & \cellcolor{lightgray}- & \cellcolor{lightgray}- \\ \midrule
\multirow{2}{*}{OVD} &  FnP & ECCV'24 & 20.25 & 11.11 & 0.40 & \cellcolor{lightgray}10.59 & 9.19 & 0.11 & 0.71 & 0.62 & \cellcolor{lightgray}2.66 & \cellcolor{lightgray}6.06 \\
 
& \textbf{Ours-OVD} & - & 42.67 & 22.39 & \textbf{1.69} & \cellcolor{lightgray}\textbf{22.25} & \textbf{22.69} & \textbf{1.45} & \textbf{7.77} & \textbf{1.15} & \cellcolor{lightgray}\textbf{8.26} & \cellcolor{lightgray}\textbf{14.26} \\ \bottomrule
\end{tabular}%
}
\end{table*}

\subsection{Experimental Results} 

Tables \ref{tab:app_vn_NtoK} to \ref{tab:app_vr_KtoArgo2} show more detailed performance among all classes. As shown in them, across all GCFS tasks, compared to existing methods, our method achieves more accurate object detection performance for overall common and novel classes, especially for ``Car'', ``Truck'', ``Van'', ``Cyclist'', ``Tram'', ``Utility\_Vehicle'', and ``Bus''. It demonstrates our method's strong knowledge transferability from the common object in the source domain while effectively generalizing to novel classes with the few-shot samples.
Regarding the evaluation on high-density (64-beam) KITTI or low-density (16-beam) A2D2 or (32-beam) Argoverse 2, the superior performance of our method underscores its strong adaptability to both moderate and extreme domain shifts. Especially in KITTI $\rightarrow5$shot-A2D2 GCFS task (Tables \ref{tab:app_vn_KtoA} and \ref{tab:app_pn_KtoA}), the performance of our method surpasses the second-best performance significantly (i.e., overall mAP: VoxelRCNN $2.74\%\rightarrow6.5\%$ and PVRCNN++ $3.17\%\rightarrow7.76\%$), ensuring robust few-shot detection even in challenging low-density scenarios. Also, as shown in Table~\ref{tab:app_vr_KtoArgo2}, where an extreme semantic shift exists, our proposed method shows the highest overall performance, indicating its superiority on fast adaptation to novel semantics under minimal target supervision.
The results of indoor 3D FSL methods (i.e., Proto-Vote, PVAE-Vote, CP-Vote) reflect the challenges in extending them to outdoor scenarios that are characterized by sparse point clouds at greater distances, dynamic objects, and varying lighting and weather conditions. Especially for common classes shared between source data and target data, those methods struggle with the demands of outdoor environments, resulting in reduced accuracy and robustness in the outdoor detection contexts.
GFS-Det performs well mostly in common classes, especially on ``Pedestrian'' objects, indicating that its dedicated category-specific branches reduce the interference of novel objects to common objects well learned in source pre-training. Yet, this separate-branch learning strategy forces the novel-object branch to learn geometric features from scratch, preventing it from leveraging geometric priors from common classes like cars or pedestrians. As a consequence, GFS-Det struggles with novel classes, hindering its ability to generalize effectively to newly learned objects and limiting its adaptability in GCFS tasks.

Table~\ref{tab:appendix_component_ablation} shows the performance of our proposed image-guided multi-modal fusion (denoted as Image-Fusion) method and our proposed contrastive-learning-enhanced prototype learning (denoted as CL-Proto) among all object classes. The experimental results show that our method demonstrates significant advantages in both novel and common classes, especially when combining Image-Fusion and CL-Proto, as in row (d), where it achieves the best performance. Specifically, the introduction of Image-Fusion significantly improves the performance on novel classes, raising the mAP from 5.48\% to 11.17\%. This improvement is particularly evident in classes like ``Van'', ``Cyclist'', and ``Tram'', where data scarcity makes single-modal features insufficient. Leveraging image-guided multi-modal fusion enables the model to better capture features in novel classes, enhancing adaptability in few-shot scenarios.
On the other hand, our proposed contrastive-learning-enhanced prototype learning mainly enhances the performance on common classes. When CL-Proto is added alone, the mAP for common classes increases from 12.77\% to 14.80\%, with a particularly notable improvement in the ``Car'' class, where mAP rises from 30.08\% to 37.07\%. Our proposed contrastive-learning-enhanced prototype learning improves the detection model with intra-class and inter-class differentiation, allowing the model to more accurately identify various features against source and target domain gaps.
When Image-Fusion and CL-Proto are combined, as in row (d), the model achieves optimal performance in both novel and common classes, with an overall mAP reaching 13.55\%. For novel classes, the mAP increases to 11.72\%, and for common classes, it rises to 15.99\%. This combination fully leverages the multi-modal feature representation strengths of our proposed image-guided multi-modal fusion method and our proposed contrastive-learning-enhanced prototype learning, enabling the model to perform better in the GCFS task. 
Notably, through mixed-precision VLM acceleration, optimizations to the model's pre- and post-processing, and other engineering improvements, our method achieves 10.11 FPS on the NVIDIA A100 GPU in the representative NuScenes→KITTI setting.

Table~\ref{tab:appendix_K_shots} shows the performance of our method under different $K$-shot target data. The results show that as the number of few-shot samples K increases, the model’s overall performance improves steadily. For instance, when K increases from 1 to 40, the overall mAP rises from 3.44\% to 28.21\%, indicating that a higher sample count helps the model better learn target features and improve detection accuracy. 
This trend suggests that with more samples, the model can effectively learn features for categories with abundant data.
Regarding the ``Van'', ``Person\_sitting'', and ``Tram'' categories, performance exhibits irregular fluctuations as $K$ increases (i.e. $5\rightarrow10$). This variation may stem from the randomness in frame sampling for few-shot conditions. Given the limited frames, the quality of each object can vary, affecting the model’s stability and consistency.
Additionally, \textit{Full-shot} training results indicate that even with training on the entire dataset, certain categories such as ``Truck'', ``Person\_sitting'', ``Cyclist'', and ``Tram'' show relatively low detection accuracy. On one hand, the limited quantity of some categories (488 trucks, 224 trams, and 56 sitting persons w.r.t. 3769 training frames) in the training data restricts the model’s ability to fully learn their features, resulting in lower accuracy. On the other hand, image-guided approach enhances the discovery of novel semantics, boosting recall on novel objects (e.g., ``Person\_sitting'', ``Cyclist'', and ``Tram'').

The results in Table~\ref{tab:appendix_OV_experiment} show that under the unsupervised few-shot setting, our extended OVD method demonstrates significant advantages across both common and novel classes. Although ST3D++ performs well in the \textit{\textbf{car}} class, its performance is limited for other common classes, highlighting its lack of generalization in few-shot scenarios. Meanwhile, FnP’s initial advantage is largely due to its cautious box candidate search strategy, which works effectively in traditional OVD settings by leveraging a large amount of target data to accumulate good object samples. However, this approach is inadequate for dealing with few-shot data, due to even fewer object samples for object feature learning. In contrast, our method achieves overall mAPs of 22.25\% on common classes and 8.26\% on novel classes, with a notable mAP boost on ``Pedestrian'', ``Van'', and ``Cyclist''. This shows that our method, under the unsupervised few-shot setting, can effectively handle feature distribution differences in the target domain, achieving more accurate and balanced detection for both common and novel classes.

\begin{table}[]
\centering
\caption{BEV/FV AP (\%) in the bird's eye view and front view of the VoxelRCNN detection under NuScenes $\rightarrow5$shot-KITTI (N→FS-K), Waymo $\rightarrow5$shot-KITTI (W→FS-K), KITTI $\rightarrow5$shot-A2D2 (K→FS-A)}
\label{tab: BEV_BBox_comp}
\resizebox{0.98\columnwidth}{!}{%
\begin{tabular}{@{}ccccc@{}}
\toprule
\multicolumn{2}{c}{\textbf{Settings}}        & \textbf{Common}      & \textbf{Novel}       & \textbf{Overall}     \\ \midrule
\multirow{2}{*}{N→FS-K} & Target-FT & 24.16/36.16 & 6.00/7.32      & 13.79/19.68 \\
& Ours      & 29.39/40.57 & 12.7/16.37  & 19.86/26.74 \\ \midrule
\multirow{2}{*}{W→FS-K} & Target-FT & 31.89/47.12 & 13.54/15.56 & 21.4/29.08  \\
& Ours      & 33.89/48.84 & 18.53/22.23 & 25.11/33.63 \\ \midrule
\multirow{2}{*}{K→FS-A} & Target-FT & 10.1/45.92  & 0.86/4.05   & 5.48/24.99  \\
& Ours      & 16.03/61.33  & 5.73/11.03   & 10.88/36.18  \\ \bottomrule
\end{tabular}%
}
\end{table}

Table~\ref{tab: BEV_BBox_comp} presents the 2D Average Precision (AP) results of VoxelRCNN under various cross-domain 5-shot settings. Compared to the Target-FT baseline, the proposed method consistently improves performance across all scenarios, especially for novel classes and in more challenging domain shifts such as KITTI to A2D2. Notably, the proposed method significantly boosts AP in both bird's eye view (BEV) and front view (FV), demonstrating its strong generalization ability in few-shot settings. These improvements highlight the method’s effectiveness in enhancing detection for unseen categories and its robustness in handling domain discrepancies.

\end{document}